\documentclass[journal]{IEEEtran}

\usepackage[noadjust]{cite}
\usepackage{amsmath}
\usepackage{url}
\usepackage{xcolor}
\usepackage{times}
\usepackage{epsfig}
\usepackage{graphicx}
\usepackage{caption}
\usepackage{float}
\usepackage{amssymb} 
\usepackage{footmisc}
\usepackage{lineno}
\usepackage{color}
\usepackage{subfigure}
\usepackage{multirow}
\usepackage{dsfont}
\usepackage{mathtools}
\usepackage{setspace}
\usepackage{csquotes}
\usepackage{breqn}
\usepackage{lettrine}
\usepackage{upquote}
\usepackage{adjustbox}
\usepackage{wrapfig}
\usepackage{lipsum}
\usepackage{mathtools}
\usepackage{comment}
\usepackage{bm}
\usepackage{array}
\usepackage{tabu}
\usepackage{booktabs}
\usepackage{enumerate}
\DeclareCaptionFormat{myformat}{\fontsize{8}{10}\selectfont#1#2#3}
\DeclareMathAlphabet\mathbfcal{OMS}{cmsy}{b}{n}
\DeclareMathOperator*{\argmax}{argmax}
\DeclareMathAlphabet{\pazocal}{OMS}{zplm}{m}{n}
\DeclareMathAlphabet{\mathpzc}{OT1}{pzc}{m}{it}

\usepackage[ruled,vlined]{algorithm2e}
\usepackage[colorlinks=true,linkcolor=blue]{hyperref}
\def\arrvline{\hfil\kern\arraycolsep\vline\kern-\arraycolsep\hfilneg}



\ifCLASSINFOpdf
\else
\fi

\begin{document}
\bstctlcite{IEEEexample:BSTcontrol}
\title{AquaticCLIP: A Vision-Language Foundation Model for Underwater Scene Analysis}

\author{Basit Alawode, Iyyakutti Iyappan Ganapathi, Sajid Javed, Naoufel Werghi,~\IEEEmembership{IEEE Senior Member}, Mohammed Bennamoun, and Arif Mahmood
%Naoufel Werghi,~\IEEEmembership{IEEE Senior Member}
\thanks{B. Alawode, I.I. Ganapathi, S. Javed, and Naoufel Werghi are with the department of computer science, Khalifa University of Science and Technology, P.O Box: 127788, Abu Dhabi, UAE. (email: sajid.javed@ku.ac.ae).}
\thanks{A. Mahmood is with the Department of Computer Science, Information Technology University, Lahore, Pakistan.}
}
\maketitle

\begin{abstract}
%\vspace{-2mm}
The preservation of aquatic biodiversity is critical in mitigating the effects of climate change. 
Aquatic scene understanding plays a pivotal role in aiding marine scientists in their decision-making processes. 
In this paper, we introduce AquaticCLIP, a novel contrastive language-image pre-training model tailored for aquatic scene understanding. 
AquaticCLIP presents a new unsupervised learning framework that aligns images and texts in aquatic environments, enabling tasks such as segmentation, classification, detection, and object counting. 
By leveraging our large-scale underwater image-text paired dataset without the need for ground-truth annotations, our model enriches existing vision-language models in the aquatic domain. 
For this purpose, we construct a 2 million underwater image-text paired dataset using heterogeneous resources including YouTube, Netflix, NatGeo, etc.
To fine-tune AquaticCLIP, we propose a prompt-guided vision encoder that progressively aggregates patch features via learnable prompts, while a vision-guided mechanism enhances the language encoder by incorporating visual context. 
The model is optimized through a contrastive pre-training loss to  align visual and textual modalities. 
AquaticCLIP achieves notable performance improvements in zero-shot settings across multiple underwater computer vision tasks, outperforming existing methods in both robustness and interpretability. 
Our model sets a new benchmark for vision-language applications in underwater environments. 
The code and dataset for AquaticCLIP are publicly available on GitHub at xxx.
\vspace{-2mm}
\end{abstract}

\IEEEpeerreviewmaketitle

\section{Introduction}
\label{sec:intro}
Global aquatic\footnote{Aquatic is a broad term encompassing all water-based environments, including both freshwater and saltwater habitats. It encompasses a vast range of ecosystems, from rivers and lakes to oceans and coral reefs.}  ecosystems are under severe threats from human activities such as overfishing and coastal development, along with climate change impacts \cite{doney2012climate, halpern2008global, jennings1998effects, zhou2022geological}. Effective conservation efforts depend on precise monitoring, which requires an accurate and automatic aquatic scene understanding system \cite{duarte2020rebuilding,grorud2021mpa}. However, the complexity of understanding aquatic environments demands significant expertise from ocean scientists and marine biologists, creating challenges for efficient monitoring \cite{zheng2023marinegpt,saleh2022computer}.

\captionsetup{type=figure}
\begin{figure*}
\begin{center}
    \centering
    \begin{subfigure} %[b]{0.76\textwidth} % Adjust width as needed
        \centering
        \includegraphics[width=0.75\textwidth, height=7.5cm]{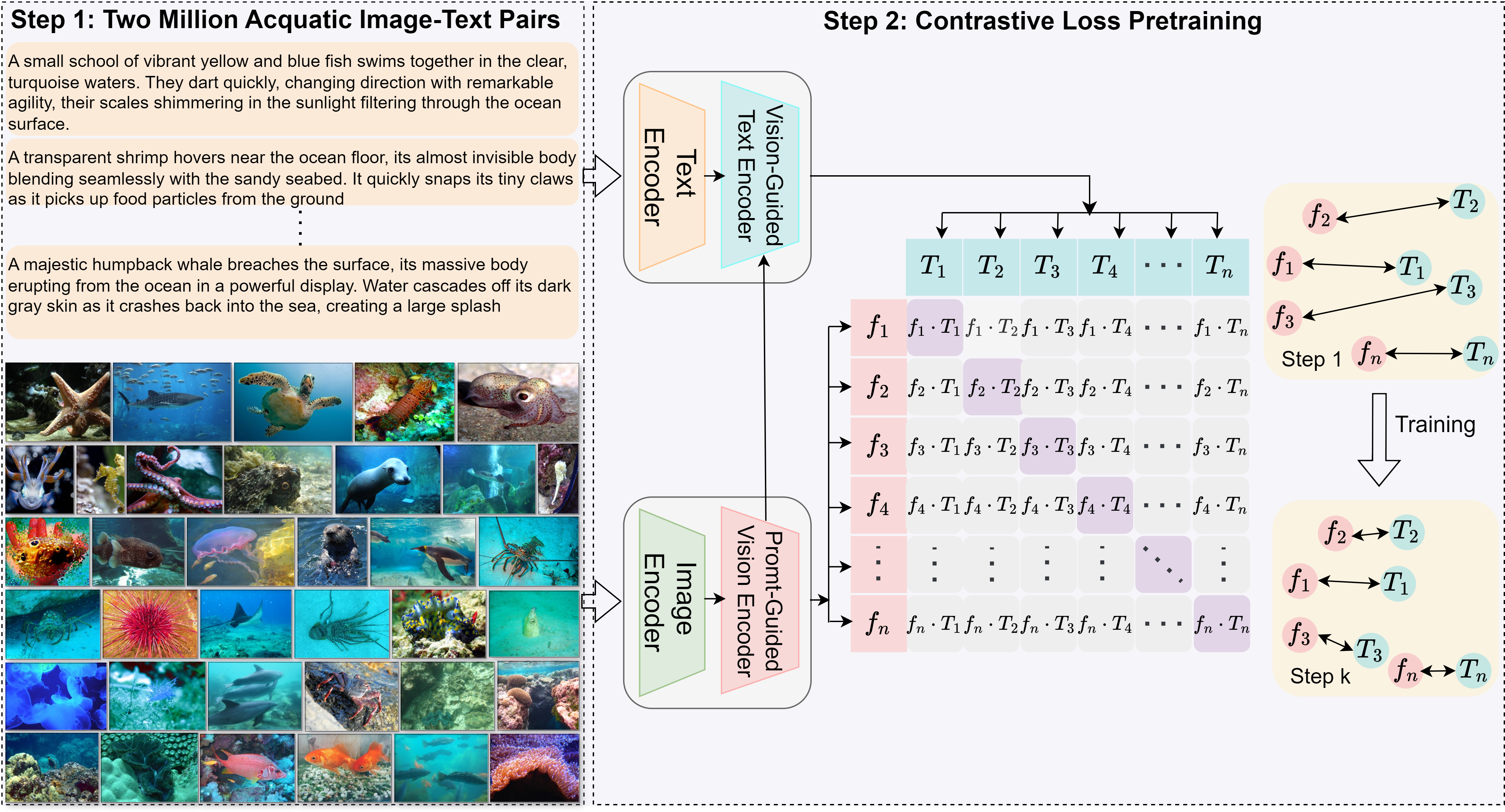} 
    \end{subfigure}      
    \begin{subfigure} %[b]{0.23\textwidth}
        \centering
        \includegraphics[width=0.23\textwidth, height=7.5cm]{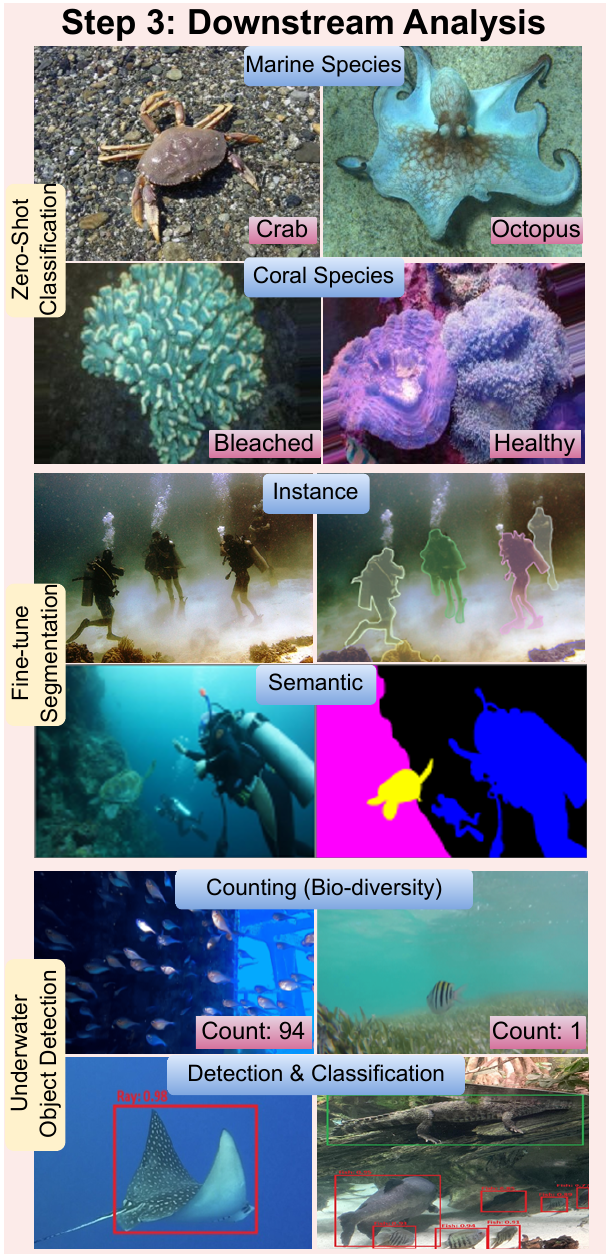}
    \end{subfigure}
     \vspace{-4mm}
     \caption{\textbf{(a) Step 1:} \textit{Two Million Aquatic Image-Text Pairs.} 
The dataset consists of paired aquatic images and enriched textual descriptions, which serve as input to the model. 
\textbf{(b) Step 2:} \textit{Contrastive Loss Pretraining}. Text and image pairs are processed by a text encoder and an image encoder.
The embeddings are aligned through contrastive loss, reducing the distance between matching pairs and improving the model's ability to associate images with their corresponding textual descriptions. \textbf{(c) Step 3:} \textit{Downstream Analysis.} AquaticCLIP performance is evaluated across various tasks such as zero-shot marine species classification, fine-tuned instance and semantic segmentation, object detection, and biodiversity counting in underwater imagery.}
\vspace{-1mm}
    \label{fig1}    
\end{center}
\end{figure*}

Recently, Vision-Language Models (VLMs) have gained increasing attention in the computer vision field \cite{radford2021learning, sammani2022nlx, mu2024embodiedgpt, maniparambil2023enhancing, yan2021videogpt}.
These models are typically pre-trained using large-scale image-text paired data, readily available online \cite{radford2021learning, singh2022flava, zhang2024vision}.
Pre-trained VLMs have been successfully applied to downstream tasks such as image classification \cite{naeem2023i2mvformer}, detection \cite{zhu2020don}, tracking \cite{li2023citetracker}, and human action recognition \cite{sun2022human}.
Their success is primarily attributed to contrastive pre-training loss, which pulls paired images and texts closer while pushing unrelated ones apart in the embedding space \cite{li2021supervision, radford2021learning}.
A notable example is Contrastive Language-Image Pre-training (CLIP), which captures rich vision-language correspondence and enables zero-shot predictions by matching the embeddings of images and texts \cite{radford2021learning,zhang2024vision}.

While significant progress has been made in extending CLIP to various computer vision tasks, few works have applied VLMs to aquatic environments \cite{zheng2024coralscop, zheng2023marinegpt,ziqiang2024marineinst}.
For example, MarineGPT, pre-trained on 5M images, was introduced for marine image-question answering tasks \cite{zheng2023marinegpt}.
More recently, MarineInst, a marine foundational model, has been proposed for segmentation and caption generation tasks \cite{ziqiang2024marineinst}.
However, the development of marine VLMs has lagged behind terrestrial VLMs due to the unique challenges of aquatic environments. 
Existing open-air datasets like COCO \cite{lin2014microsoft} cannot be used to train aquatic VLMs, and large-scale, domain-specific aquatic image-text pairs are not readily available. 
This scarcity of data makes pre-training aquatic VLMs particularly challenging.

To bridge this gap, we propose a large-scale aquatic image-text paired dataset comprising 2 million image-text pairs. 
We then introduce AquaticCLIP, a model that efficiently aligns aquatic images and texts for several downstream tasks, as shown in Fig. \ref{fig1}.
Our dataset is collected from publicly available resources such as National Geographic (NatGeo) \cite{doe2023coral, national_geographic_coral, ng_coral_bleaching, ng_ocean_acidification,ng_great_barrier_reef,ng_coral_conservation,ng_ocean_biodiversity}, aquatic biology textbooks and journals, YouTube aquatic documentary videos, Fishes of Australia \cite{van2014family, schodde1997zoological, merrick2006australasian,shelley2017revision}, Marine Twitter, Netflix, and the Corals of the World \cite{veron2016corals}.
To our knowledge, no such paired dataset exists for aquatic scenes, except MarineInst \cite{ziqiang2024marineinst}, which contains only images. 
To further enrich the textual descriptions, we  generate additional descriptions for aquatic images using MarineGPT \cite{zheng2023marinegpt} both at the image level and the instance level. 
For instance-level descriptions, we detect objects within an aquatic image using a pre-trained object detector and then MarineGPT is employed for each instance.
These additional descriptions are then refined using a custom cleaning module to remove irrelevant keywords while retaining those relevant to the aquatic imagery.

For our AquaticCLIP model, we introduced two lightweight learnable encoders for the image and text branches, respectively. 
The key idea behind these encoders is to effectively transfer the VLM into the aquatic domain by leveraging the prior knowledge of the existing MarineGPT. 
To enable the VLM to process aquatic images more efficiently, we designed a prompt-guided vision encoder based on prompt-based learning. 
Specifically, for the image branch, we aggregate all patch features using learnable weights. 
A set of learnable prompt vectors is introduced to progressively guide the fusion of patch features, grouping similar ones together. 
This method allows each prompt to capture more global contextual information for final similarity computation.

For the text branch, we propose a vision-guided text encoder that integrates corresponding image information into the text encoder. 
By employing a multi-modal text encoder for guidance, knowledge is more effectively transferred to the VLM's text encoder. 
The visual representations and textual descriptions learned by these two encoders are then aligned using a contrastive pre-training loss, similar to CLIP \cite{radford2021learning}.
The main goal is to bring similar visual and textual concepts of aquatic images closer and push dissimilar ones apart.

We conducted extensive experimental evaluations of the AquaticCLIP model on various downstream tasks, including zero-shot and fine-tuning scenarios. 
Zero-shot tasks included aquatic species recognition, fine-grained fish classification, coral species classification, and cross-modal retrieval.
Fine-tuning tasks involved coral segmentation, instance segmentation of aquatic imagery, semantic segmentation, aquatic object detection and classification, and aquatic object counting. 
Our results demonstrate significant performance improvements compared to existing State-Of-The-Art (SOTA) methods in aquatic settings.

The key contributions of this work include: 
\begin{comment}
 \begin{enumerate}
\item We propose a large-scale 2M image-text paired dataset specifically designed for pre-training aquatic VLMs (Sec. \ref{sec:dataset}).
\item \textcolor{blue}{To further enrich the textual descriptions, for the aquatic images in our dataset, we also generate pseudo-descriptions using MarineGPT in zero-shot settings at the image and instance levels (Sec. \ref{sec:generation}).
These descriptions are then refined to remove undesired keywords, improving overall data quality (Sec. \ref{sec:cleaning}).}
\item We introduce a learnable prompts-guided image encoder that enhances the encoding of aquatic images. 
By utilizing learnable prompts, irrelevant image regions are suppressed, resulting in more effective visual representations (Sec. \ref{sec:pgve}).
\item We propose a vision-guided text encoder that aligns textual descriptions with corresponding visual context more accurately (Sec. \ref{sec:vgte}). 
This refined textual representation is used for efficient pre-training of the VLM (Sec. \ref{loss}).
\item  Extensive experimental evaluations are conducted on various downstream aquatic computer vision tasks, including instance segmentation, semantic segmentation, zero-shot aquatic species recognition, aquatic object counting, and challenging aquatic object detection. 
Our results show significant performance improvements compared to existing SOTA methods in aquatic environments (Sec. \ref{sec:results}).
\end{enumerate}
\end{comment}

\begin{enumerate}
\item We propose a large-scale dataset of 2 million image-text pairs for aquatic VLM pre-training, with enriched textual descriptions generated by MarineGPT in zero-shot settings at both the image and instance levels and refined to exclude undesired keywords, enhancing data quality (Sec. \ref{sec:dataset}-\ref{sec:cleaning}).
\item We introduce a dual-encoder approach with a prompts-guided image encoder to suppress irrelevant image regions and a vision-guided text encoder for improved alignment of visual and textual representations (Sec. \ref{sec:pgve}-\ref{loss}).
\item  Extensive evaluations on diverse aquatic vision tasks show significant gains over existing SOTA methods (Sec. \ref{sec:results}).
\end{enumerate}

\noindent The remainder of this paper is organized as follows: Sec. \ref{sec:relatedwork} reviews related work. 
Sec. \ref{sec:method} presents our proposed dataset and the AquaticCLIP. 
Sec. \ref{sec:results} presents the experiments and results, and Sec. \ref{sec:conclusion} concludes our work.

\section{Related Work}
\label{sec:relatedwork}
Numerous deep learning-based approaches have been proposed for various aquatic scene understanding tasks \cite{saleh2022computer, wang2023deep, zhang2024webuot, li2023underwater, gonzalez2023survey} including underwater image enhancement  \cite{desai2022aquagan, wang2022agcyclegan, islam2020fast}, species classification \cite{li2023deep, xu2023systematic}, underwater object recognition \cite{khan2023fishnet, jalal2020fish}, coral segmentation \cite{zheng2024coralscop}, Object counting \cite{sun2023indiscernible}, and underwater tracking \cite{zhang2024webuot}.
%While CNNs have traditionally been used, more recent approaches have incorporated vision transformers.
%Below, we discuss key categories of methods used in specific aquatic computer vision applications.

\noindent \textbf{1. CNN-based Methods:} %CNNs have gained popularity due to their ability to automatically learn and extract features from aquatic images, improving accuracy in various tasks.
For underwater image enhancement and restoration, many CNN-based approaches have been proposed including WaterGAN \cite{li2017watergan}, Image-2-Image Translation \cite{cho2020underwater}, AquaGAN \cite{desai2022aquagan}, CycleGAN \cite{lu2019multi}, AGCycleGAN \cite{wang2022agcyclegan} and others \cite{hu2018underwater}.
%Hu \textit{et al.} proposed a cross-layer multi-scale CNN using separate networks for ambient light and blue channel transmission estimation \cite{hu2018underwater}.
%Li \textit{et al.} introduced WaterGAN, a model that performs color correction by estimating attenuation, backscattering, and camera characteristics \cite{li2017watergan}.
%Cho \textit{et al.} utilized image-to-image translation via GANs for image correction and enhancement \cite{cho2020underwater}, while Desai \textit{et al.} presented AquaGAN, which estimates attenuation coefficients using a UNet CNN based on the image formation model \cite{desai2022aquagan}.
%Other notable approaches include, Lu \textit{et al.} CycleGAN model combined with the dark channel prior for underwater image restoration \cite{lu2019multi} and 
%Wang \textit{et al.} AGCycleGAN for restoration \cite{wang2022agcyclegan}.
For underwater object detection and classification, several methods and datasets have led to significant progress \cite{khan2023fishnet, yeh2021lightweight, jalal2020fish}. 
Khan \textit{et al.} introduced FishNet, a benchmark dataset for fish recognition and functional trait prediction \cite{khan2023fishnet}.
%Yeh \textit{et al.} proposed a lightweight deep network for simultaneous object detection and color conversion \cite{yeh2021lightweight}, while Jalal \textit{et al.} applied deep networks with %temporal information for fish detection and species classification \cite{jalal2020fish}.

Species classification is crucial for underwater monitoring and conservation and a number of methods have been proposed in this directions \cite{10612606, Pedersen_2019_CVPR_Workshops, siddiqui2018automatic}.  
%Prathima \textit{et al.} employed models like Yolo, FRCNN, and RetinaNet for marine animal detection \cite{10612606}. 
%Pedersen \textit{et al.} released a dataset for species classification including fish, crabs, and starfish in varying visibility conditions \cite{Pedersen_2019_CVPR_Workshops}. 
%Siddiqui \textit{et al.} used a cross-layer pooling method with a pre-trained CNN for generalized species classification \cite{siddiqui2018automatic}.
Coral reefs, known as the ``rainforests of the sea'', play a critical role in biodiversity. 
However, coral segmentation is a challenging task due to complex underwater visual conditions.
Many methods are recently proposed for this task \cite{zhong2023combining, ziqiang2023coralvos}.
%Zhong \textit{et al.} developed 3D models and performed millimeter-accurate rugosity evaluation of coral habitats \cite{zhong2023combining}, while Ziqiang \textit{et al.} introduced a new dataset %and benchmark for coral video object segmentation \cite{ziqiang2023coralvos}..
Object counting is essential for marine biology, fisheries management, and environmental monitoring. 
Most approaches focus on fish and coral counting such as \cite{burguera2024deep, babu2023computer, modasshir2018coral}. 
%Burguera \textit{et al.} proposed a CNN-based method for lobster counting in deep sea environments \cite{burguera2024deep}, and Babu \textit{et al.} introduced a juvenile fish counting model using object detection \cite{babu2023computer}.
%Other approaches include simultaneous fish detection and counting \cite{babu2023computer} and coral identification and counting \cite{modasshir2018coral}.
Underwater visual tracking has also seen significant advances using CNN-based trackers and new datasets \cite{9499961, cai2023semi, hao2022umotma, lee2024detection}.
%Panetta \textit{et al.} proposed the UOT100 underwater tracking dataset and a GAN-based tracker \cite{9499961, cai2023semi, hao2022umotma, lee2024detection}.
%Cai \textit{et al.} developed a semi-supervised visual tracker for marine animals using autonomous underwater vehicles \cite{cai2023semi}, and Hao \textit{et al.} introduced a memory aggregation network for multi-object tracking \cite{hao2022umotma}.
%Further developments in underwater tracking can be found in \cite{lee2024detection}.

\noindent \textbf{2. Transformer-based Methods:} Many ViT-based techniques for detection, segmentation, counting, tracking, and classification have been proposed for underwater scenes \cite{peng2023u, zhang2024webuot, sun2023indiscernible, yang2024density, ai2024novel}.
For instance, UShape transformer \cite{peng2023u} and transformer-driven GAN \cite{ummar2023window} have been proposed for underwater image restoration tasks.
%Peng \textit{et al.} introduced the UShape transformer architecture for underwater image enhancement \cite{peng2023u}, while Mehnaz \textit{et al.} proposed a transformer-driven GAN for underwater image %restoration \cite{ummar2023window}.
Alawode \textit{et al.} proposed a combined approach for underwater image enhancement and visual tracking \cite{alawode2022utb180}, and Zhang \textit{et al.}  developed a large-scale benchmark for advancing underwater object tracking \cite{zhang2024webuot}.
A dense object counter \cite{sun2023indiscernible} and density-guided attention \cite{yang2024density} methods are proposed for underwater object counting task.
%Sun \textit{et al.} introduced a dense object counter designed for underwater scenes , and Yang \textit{et al.} developed a density-guided temporal attention transformer for underwater object counting .
For fish detection and classification \cite{liu2024dp, liu2024cffi}, and coral reef classification \cite{ai2024novel, shao2024deep} tasks, ViT-based models have also recently been proposed. 
%Liu \textit{et al.} presented a dual-path pyramid vision transformer network for fish detection \cite{liu2024dp}, as well as an enhanced vision transformer architecture for fish classification in aquaculture %\cite{liu2024cffi}.
%Bo \textit{et al.} proposed a vision transformer-based method for coral reef classification \cite{ai2024novel}, while Shao \textit{et al.} applied transformers for multi-label classification of coral %conditions \cite{shao2024deep}. 
A token-based selective ViT \cite{si2023token} and cascaded attention \cite{zhang2024catnet} models are proposed for fine-grained marine species classification.
%Si \textit{et al.} introduced a token-based selective transformer for fine-grained marine organism classification \cite{si2023token}, and Zhang \textit{et al.} proposed a cascaded attention mechanism for %marine species classification \cite{zhang2024catnet}.
Additionally, some self-supervised learning methods have also been utilized for underwater image analysis \cite{huang2023contrastive, saleh2022transformer}.\\
\noindent \textbf{3. Vision-Language Models (VLMs):} In the context of safeguarding aquatic biodiversity, there is an increasing need for VLMs to facilitate AI-based aquatic scene understanding systems. 
However, VLMs have only been sparsely applied to aquatic scene analysis \cite{zheng2024coralscop, zheng2023marinegpt,ziqiang2024marineinst}.
For instance, Zheng \textit{et al.} introduced CoralSCOP, a foundational model for the automatic segmentation of coral reefs \cite{zheng2024coralscop}, and MarineGPT, a multimodal large language model for marine object classification and question-answering \cite{zheng2023marinegpt}.
Zheng \textit{et al.} further evaluated GPT-4V for marine images, but found its performance unsatisfactory for domain-specific needs of marine biologists \cite{zheng2024exploring}. 
Recently, the MarineInst foundational model was proposed, pre-trained on 2M images for segmenting and captioning marine imagery \cite{ziqiang2024marineinst}.
To the best of our knowledge, VLMs have not been thoroughly explored for aquatic scene understanding, except for MarineGPT and MarineInst. 
Our work is the first to introduce AquaticCLIP, with comprehensive analysis and comparisons to existing SOTA methods.
%\vspace{-13mm}

\begin{figure*}[t!]
    \centering    
    \includegraphics[width=0.96\linewidth]{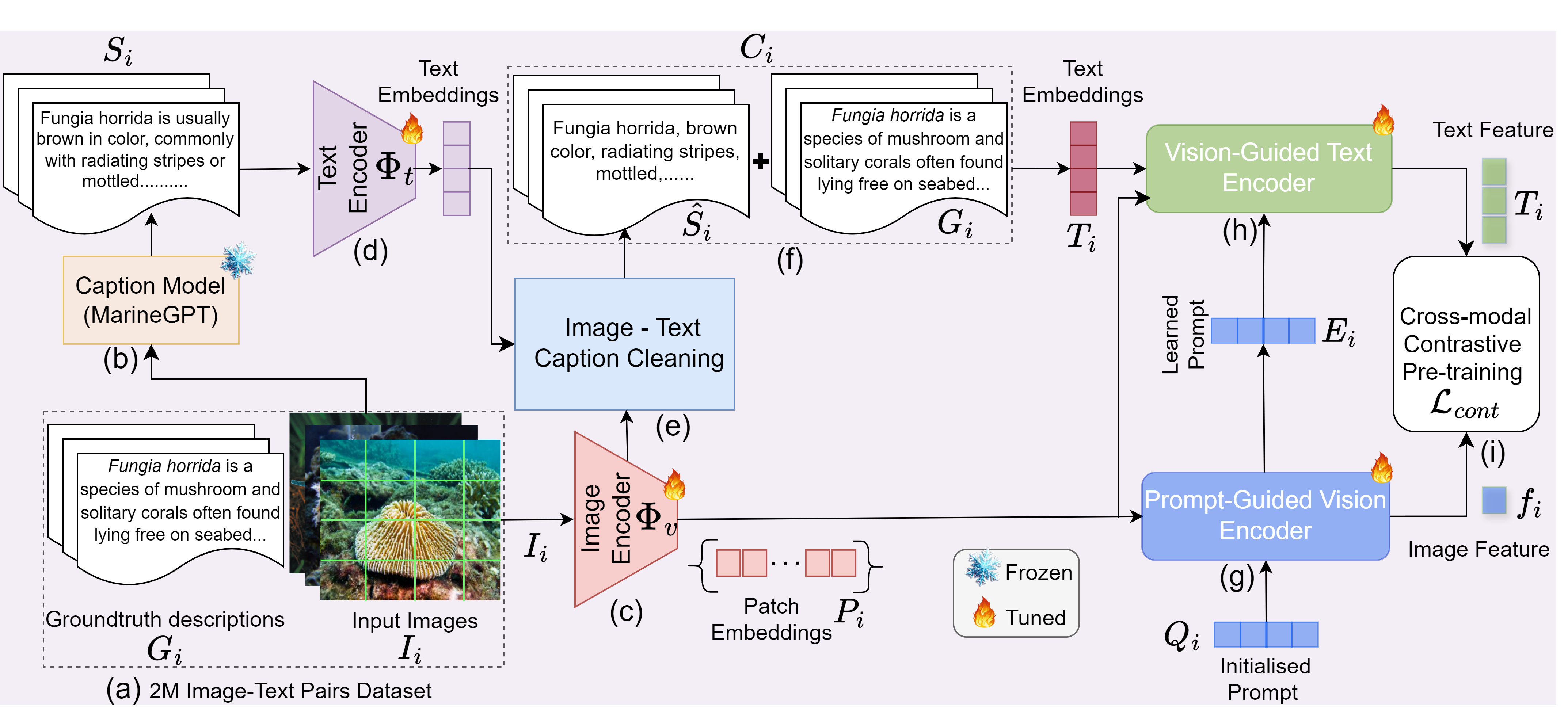}
    \vspace{-2mm}
    \caption{\textbf{Overview of AquaticCLIP architecture and training process}. 
\textbf{(a)} Shows a set of input image-text pairs. 
\textbf{(b)} A caption model (MarineGPT) generates textual descriptions for the images. 
\textbf{(c)} Input images are divided into patches and processed by the image encoder $\Phi_v$ to produce patch embeddings $\textbf{P}_{i}$.
\textbf{(d)} The generated textual descriptions $\textbf{S}_{i}$ are processed by the text encoder $\Phi_{t}$ to produce text embedings. 
\textbf{(e)-(f)} The textual description $\textbf{S}_{i}$ is then cleaned by an image-text caption cleaning module to produce refined descriptions $\hat{\textbf{S}}_{i}$ which are then combined with groundtruth descriptions $\textbf{G}_{i}$ to produce enriched textual description data $\textbf{C}_{i}$ .  
Both image and text embeddings are refined using \textbf{(h)} vision-guided text encoding and \textbf{(g)} prompt-guided vision encoding. 
The learned prompts $\textbf{E}_i$ guide the fusion of patch embeddings, while initialized prompts $\textbf{Q}_{i}$ are used to enhance the visual representation. 
\textbf{(i)} The final image and text features are aligned using a cross-modal contrastive pre-training loss $\mathcal{L}_{cont}$, ensuring a stronger association between text and image representations. }\label{introFigure}
%\vspace{-3mm}
\end{figure*}

\section{Proposed AquaticCLIP Model}
\label{sec:method}
This work introduces the Aquatic Contrastive Language-Image Pre-training (AquaticCLIP) model, which leverages a collection of 2M aquatic image-text pairs curated form several heterogeneous resources. 
The ground truth descriptions are also enriched by harnessing the existing VLM MarineGPT.
The primary objective is to pre-train AquaticCLIP using diverse aquatic data sourced from various platforms, enhancing its capability for zero-shot transfer across different aquatic imagery tasks.
This is particularly beneficial for recognizing unfamiliar marine species and coral reef categories that were not encountered during training.
Fig. \ref{introFigure} illustrates the key components of the proposed model, which include the construction of the 2M aquatic image-text paired dataset further enriched by unsupervised generated textual descriptions, caption cleaning, a lightweight prompt-guided vision encoder, a vision-guided text encoder, and the pre-training process. 
The contrastive learning approach aligns positive image-text pairs while separating negative ones. 
The details of these processes are discussed in the subsequent sections.

\subsection{Aquatic Dataset Construction and Curation}
\label{sec:dataset}
Our dataset construction pipeline involves two main steps: gathering image-text paired data from multiple sources and cleaning and filtering the collected data.
We assembled a dataset of 2 million aquatic image-text pairs from various resources, including YouTube videos, marine and ocean sciences textbooks and articles, the Corals of the World \cite{veron2016corals}, Fishes of Australia \cite{van2014family, schodde1997zoological, merrick2006australasian,shelley2017revision}, Marine Twitter, Netflix, and National Geographic (NatGeo) \cite{doe2023coral, national_geographic_coral, ng_coral_bleaching, ng_ocean_acidification,ng_great_barrier_reef,ng_coral_conservation,ng_ocean_biodiversity}.

For YouTube videos, we searched using keywords such as ``underwater world'', ``marine documentary'',``deep oceans'', ``great barrier reef'', ``aquatic scenes'', and ``coral reefs'' etc. 
For Netflix videos, we explored hundreds of documentaries, including ``My Octopus Teacher'', ``Last Breath'', and ``Wonders of the Reef'', etc.
Subtitles provided by both resources were used to generate aligned image-text pairs, which were manually checked and refined. 
Unique frames were extracted every 50 seconds from the videos, which often contained challenges like low visibility, motion blur, background clutter, and color distortions.

Additionally, We utilized 1200 diverse textbooks on marine biology and oceanography, along with research articles from ocean and marine journals and NatGeo magazines, to further enrich the dataset. 
Figures and captions were extracted using PDF-Figures 2.0 tool \cite{clark2016pdffigures}, and we manually refined the data to ensure the selected images had meaningful associated text. 
Images not related to aquatic environments were discarded.

We also included image-text pairs from the Corals of the World repository and Fishes of Australia, only selecting pairs with detailed descriptions. 
Furthermore, we used the Twitter platform to search for relevant content using hashtags like $\#$MarineBiology, $\#$Oceans, and $\#$Fisheries, considering only channels with over 100 followers. 
After a thorough cleaning and filtering process, we retained 2 million high-quality image-text pairs representing a diverse range of aquatic scenes.
\textit{More details are provided in the supplementary material.}
%%\vspace{-5mm}
\subsection{Unsupervised Generation of Image and Instance-Level Descriptions (Fig. \ref{introFigure} (b))} 
\label{sec:generation}
%%\vspace{-2mm}
In order to enrich the ground-truth textual descriptions, we generated additional textual descriptions at both the image and instance levels using a VLM MarineGPT \cite{zheng2023marinegpt}, which includes a frozen ViT image encoder and Q-former.
%For each image in our dataset, we manually verified ground truth textual descriptions, which were then combined with the pseudo-textual descriptions during the pre-training stage. 
%\textit{Our AquaticCLIP model is pre-trained on 2M image-text pairs, where the ground-truth descriptions of the images were combined with pseudo-textual descriptions.}
At the image level, each image $\textbf{I}_{i}$ is input into MarineGPT to generate its corresponding textual descriptions. 

For the instance level, we pre-trained MRegionCLIP, an object detector based on RegionCLIP \cite{zhong2022regionclip} and MarineDet \cite{haixin2023marinedet} and applied it to our 2M imagery dataset to detect all instances in zero-shot settings. 
Each instance was then passed through MarineGPT to generate a textual description.
Specifically, we used the following prompt template: ``The image is $<\textbf{image}>$. Describe the object in this image:'', where $<\textbf{image}>$ is the image token.
The generated textual descriptions $\textbf{S}_{i}$ at the image and instance levels were then cleaned using our cleaning module, which is explained in the following subsection.
%were combined with ground-truth descriptions $\textbf{G}_{i}$ to generate more enriched and comprehensive textual data $\textbf{C}_{i}$ for further processing. 
%The generated textual descriptions $\textbf{S}_{i}$ were cleaned using our cleaning module, which is explained in the following section.

\subsection{Semantic Filtering and Cleaning of Generated Textual Descriptions (Figs. \ref{introFigure} (c)-(f))}
\label{sec:cleaning}
The generated textual descriptions $\textbf{S}_{i}$ may contain noise, such as broken sentences, incorrect descriptions, or irrelevant keywords. 
To address these issues, we developed a textual description cleaning module aimed at identifying the semantically closest and most relevant keywords.

In this process, each generated textual description $\textbf{S}_{i}$ is broken down into a set of $k$-keywords ($\{\textbf{s}_{i}^{j}\}_{j=1}^{k}$).
For each keyword, we compute its cosine similarity with the image embedding as follows:
%%\vspace{-2mm}
\begin{equation}
\hat{\textbf{S}}_{i}= \argmax_{s\in S}< \Phi_{v}(\textbf{I}_{i})\cdot  \Phi_{t}(\textbf{s}_{i}^{j})>, 
\label{eqn1}
%%\vspace{-3mm}
\end{equation}

\noindent where $\Phi_{v}$ is a vision encoder followed by an MLP, and $\Phi_{t}$ is a text encoder from the CLIP model. 
We retain the top-p$\%$ of keywords in $\hat{\textbf{S}}_{i}$, discarding the rest as noise. 
This cleaning process ensures that the remaining keywords are semantically aligned with the visual content, improving the quality of the textual descriptions.

For each image in our dataset, we manually verified ground truth textual descriptions $\textbf{G}_{i}$. 
During the pre-training stage, the refined keywords $\hat{\textbf{S}}_{i}$ both at the image and instance levels were combined with ground-truth descriptions $\textbf{G}_{i}$ to generate more enriched and comprehensive textual data $\textbf{C}_{i}$ for further processing. 
\textit{Our AquaticCLIP model is pre-trained on 2M image-text pairs, where the ground-truth descriptions of the images were combined with refined keywords.}

\subsection{Prompt-guided Vision Encoder (Fig. \ref{encoder} (a))}
\label{sec:pgve}
To generate efficient visual embeddings for each aquatic image $\textbf{I}_{i}$, we aggregate the patch features of the input image using learned visual prompts. 
First, the input image $\textbf{I}_{i}$ is divided into $n_{p}$ non-overlapping patches $\{w_{i}^{j}\}_{j=1}^{n_{p}}$, each of size $m \times m$.
These patches are fed into the pre-trained image encoder $\Phi_{v}$ to produce embeddings $\textbf{P}_i=\{\textbf{p}_{i}^{j}\}\in \mathbb{R}^{d_{p} \times n_p}$.

To effectively aggregate these patch embeddings into a final image-level embedding for similarity calculation, we designed a prompt-guided image encoder, as illustrated in Fig. \ref{encoder} (a).
We randomly initialize a set of learnable prompt features $\textbf{Q}_{i}=\{r_{i}^{j}\}_{j=1}^{n_{r}} \in \mathbb{R}^{d_p \times n_r}$, where $n_{r}$ represents the number of learnable prompts. 
These prompts guide the progressive fusion of patch embeddings. 
Cross-attention is then computed using the visual embeddings as keys $\textbf{K}_{i}=\textbf{P}_i$ and values $\textbf{V}_{i}=\textbf{P}_i$, while the prompts $\textbf{Q}_{i}$ serve as queries.
%%\vspace{-5mm}
\begin{equation}
\textbf{E}_{i}=\textrm{Softmax} \Bigg( \frac{\textbf{Q}_{i}\textbf{K}_{i}^\top}{\sqrt{d_{p}}}\Bigg)\textbf{V}_{i},\\
\textbf{E}_{i}=\textrm{Norm}(\textbf{E}_{i})+\textbf{Q}_i,
\label{eqn2}
%%\vspace{-3mm}
\end{equation}
\noindent The learnable prompts help prioritize patches with high semantic similarity, resulting in a more meaningful image-level representation that captures global contextual information. 
The final image-level features are derived using an attention-based feature fusion method, as shown below:
%%\vspace{-3mm}
\begin{equation}
\textbf{E}^{'}_{i}=\textbf{W}_{1}\textbf{E}_{i},\textbf{e}^{}_{i}=\exp(\textbf{W}_{3}^{\top}(\textrm{tanh}(\textbf{W}_{2}\textbf{E}^{'}_{i}))),
%%\vspace{-1mm}
\end{equation}

\noindent Here, $\textbf{W}_{1}$, $\textbf{W}_{2} \in \mathbb{R}^{d_p\times d_p}$ and $\textbf{W}_{3 }\in \mathbb{R}^{1\times d_p}$ are learnable matrices, and the softmax function is used to compute attention weights  $\textbf{a}_i(j)$. 
The image-level representation $\textbf{f}_{i}$ is then computed as follows:
%%\vspace{-3mm}
\begin{equation}
\textbf{a}_i(j)=\frac{\textbf{e}_i(j)}{\sum_{k=1}^{n_r} \textbf{e}_i(k)},~\textrm{and}~\textbf{f}_{i}=\textbf{W}_{4} \sum_{j=1}^{n_{r}}\textbf{a}_i(j)\textbf{E}^{'}_{i}(j),
%%\vspace{-4mm}
\end{equation}
\noindent where $\textbf{W}_{4}\in \mathbb{R}^{d_{p} \times d_p}$ is a learnable weight matrix and $\textbf{E}^{'}_{i}(j)$ is a column vector of $\textbf{E}^{'}_{i}$.

\begin{figure}[t!]
    \centering    
    \includegraphics[width=3.5in, height=2.8in]{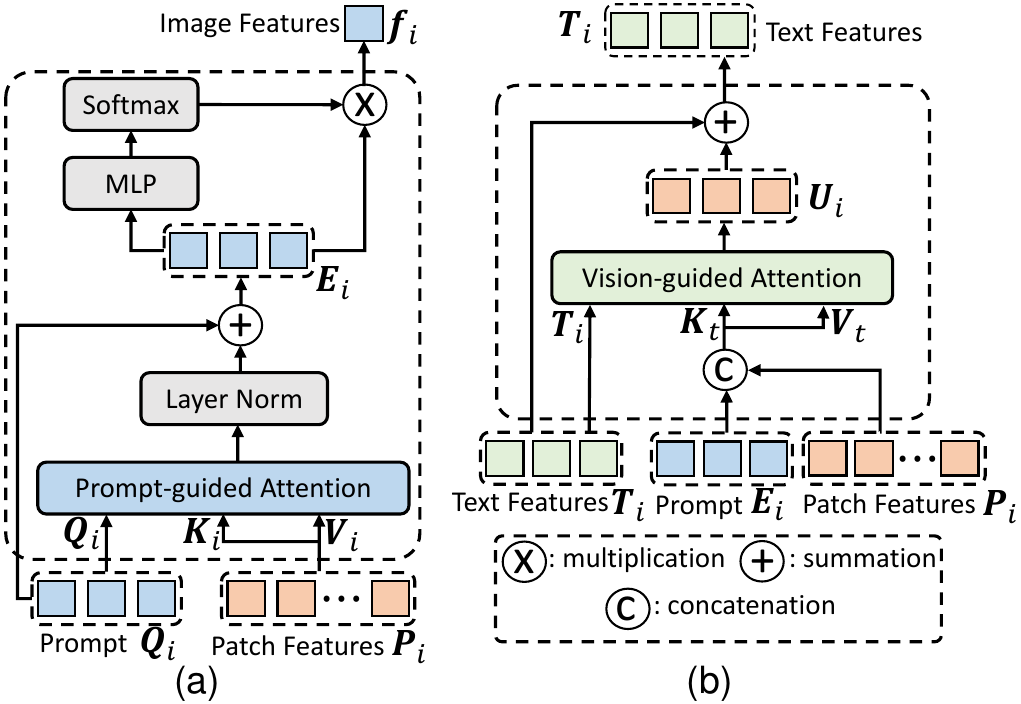}
    %%%\vspace{-8mm}
    \caption{\textbf{(a) Prompt-Guided Vision Encoder:} The prompt-guided attention mechanism combines patch features $\textbf{P}_{i}$ with initialized prompts $\textbf{Q}_{i}$ through layer normalization and an MLP, followed by softmax to produce the final image features $\textbf{f}_{i}$.
\textbf{(b) Vision-Guided Text Encoder:} Text embeddings $\textbf{T}_{i}$ are refined using a vision-guided attention mechanism, where patch features $\textbf{P}_{i}$, learned prompts $\textbf{E}_{i}$, and text embeddings $\textbf{T}_{i}$ are concatenated to compute attention $\textbf{U}_{i}$, which further enhances $\textbf{T}_{i}$.}
    \label{encoder}
    %%\vspace{-10mm}
    \end{figure}

\subsection{Vision-guided Text Encoder (Fig. \ref{encoder} (b))}
\label{sec:vgte}
In the text encoder branch, the enriched textual descriptions $\textbf{C}_{i}$ are fed into the CLIP text encoder to obtain textual representations $\textbf{T}_{i}$  corresponding to the descriptions of the $i$-th image. 
These representations are then passed through a vision-guided attention layer for refinement. 
The patch features $\textbf{P}_{i}$ and the learned prompts $\textbf{E}_{i}$ are concatenated as $\textbf{V}_i$, which serves as the key $\textbf{K}_{t}$ and value $\textbf{V}_{t}$, while the textual representations $\textbf{T}_{i}$ are used as the query, as shown in (Fig. \ref{encoder} (b)).
The vision-guided attention mechanism is computed as follows:
%%\vspace{-5mm}
\begin{equation}
\textbf{U}_{i}=\textrm{Softmax} \Bigg( \frac{\textbf{T}_{i}\textbf{K}_{t,i}^\top}{\sqrt{d_{p}}}\Bigg)\textbf{V}_{t,i},\\
\textbf{T}_{i}=\textbf{T}_{i}+\textbf{U}_i,
\label{eqn}
%%\vspace{-2mm}
\end{equation}
\noindent This context-guided text encoder further refines the textual features by incorporating image context and learned visual prompts.
This process enhances the alignment between images and texts, improving the performance of the AquaticCLIP model.
%%%\vspace{-1mm}

\subsection{Cross-Modal Contrastive Loss for Vision-Language Alignment (Fig. \ref{introFigure} (i))}
\label{loss}
We pre-train our prompt-guided vision encoder and vision-guided text encoder using a cross-modal contrastive loss function. 
This loss is formulated as a temperature-scaled vision-language pre-training loss, similar to $W$-way classification, where $W$ represents the batch size of image-text pairs involved in the training process \cite{radford2021learning, chen2020simple, tian2020contrastive}.
Given a batch of $W$ paired normalized image and text embeddings $\{\textbf{f}_{i},\textbf{T}_{i}\}$, we minimize the contrastive loss in two directions: image-to-text ($i \rightarrow t$) and text-to-image ($t \rightarrow i$) as:
%The image-to-text loss is defined as:
%%\vspace{-2mm}
\begin{equation}
\mathcal{L}_{i2t}= -\frac{1}{W}\sum_{i=1}^{W} \log \frac{\exp (\tau \textbf{T}_{i}^{\top}\textbf{f}_{i} )}{\sum_{j=1}^{W}\exp(\tau \textbf{T}_{i}^{\top}\textbf{f}_{j})},
%%%\vspace{-2mm}
\end{equation}
%\noindent The text-to-image loss is defined as:
%%\vspace{-2mm}
\begin{equation}
\mathcal{L}_{t2i}=-\frac{1}{W}\sum_{j=1}^{W} \log \frac{\exp (\tau \textbf{f}_{j}^{\top}\textbf{T}_{j} )}{\sum_{i=1}^{W}\exp(\tau \textbf{f}_{j}^{\top}\textbf{T}_{i})},  
%%\vspace{-2mm}
\end{equation}

\noindent where $\tau$ is a learnable temperature parameter that controls the smoothness of the distribution \cite{radford2021learning}.
The overall contrastive loss, $\mathcal{L}_{cont}$ , is the sum of both losses: $\mathcal{L}_{cont}=\mathcal{L}_{i2t}+\mathcal{L}_{t2i}$.
This loss minimizes the distance between embeddings of positive image-text pairs and maximizes the distance between negative pairs $(\textbf{f}_{i},\textbf{T}_{j})$, where $i\ne j$, ensuring that images and texts with the same semantic content have similar representations in the feature space.

\subsection{Zero-shot Transfer for Image Classification}
%%\vspace{-2mm}
Radford \textit{et al.} introduced the concept of zero-shot classification using a prompt-based approach \cite{radford2021learning}.
In our method, each class in the test dataset is converted into one or more text prompts using predefined templates, such as ``An image of \{Sea Urchins\}." or ``An image of \{Oyster\}." 
For each test image, we compute the $\ell_{2}$ normalized embeddings using our prompt-guided vision encoder and vision-guided text encoder. 
We then calculate the cosine similarity between the test image and the set of testing prompts to find the best match, resulting in zero-shot classification. 
Additional details are provided in the supplementary material.
%%\vspace{-4mm}

\section{Experimental Evaluations}
\label{sec:results}
%\vspace{-2mm}
We conducted extensive experiments to evaluate the performance of the proposed AquaticCLIP model across various tasks, including zero-shot classification of marine species, fine-grained fish, and coral species classification. 
Additionally, we performed zero-shot cross-modal retrieval tasks for aquatic images (\textit{see supplementary material}). 
For downstream tasks, we applied fine-tuned instance segmentation, semantic segmentation of underwater imagery, as well as marine object detection, classification, and counting (\textit{see supplementary material}). 
These experiments, covering a range of classification, detection, and segmentation tasks, allowed us to thoroughly assess AquaticCLIP's performance. 
We also compared our results with SOTA methods, including both VLMs and vision-only approaches.

\begin{table*}[t!]
\caption{\textbf{Architectural ablation} highlighting the importance of the Prompt-Guided Vision Encoder (PGVE) and Vision-Guided Text Encoder (VGTE), as well as the impact of frozen versus fine-tuned image $\Phi_{v}$ and text encoders $\Phi_{t}$.
The results, reported as $F_{1}$ scores, reflect zero-shot classification performance on six datasets.
The fully fine-tuned AquaticCLIP model with both PGVE and VGTE achieves the highest scores, demonstrating the effectiveness of fine-tuning and these key components.}
%\vspace{-7mm}
\begin{center}
\makebox[\linewidth]{
\scalebox{0.90}{
\begin{tabu}{|[2pt]c|c|c|c|c|[2pt]c|c|c|c|c|c|[2pt]}
\tabucline[0.5pt]{-}
Variants&Frozen ($\Phi_{v}$)&Frozen ($\Phi_{t}$)&PGVE &VGTE&MAI&SAI&FishNet&FNOI&LSF&CC\\\tabucline[0.5pt]{-}
Frozen CLIP&\checkmark&\checkmark&$\times$&$\times$&0.692&0.702&0.651&0.622&0.772&0.752\\\tabucline[0.5pt]{-}
AquaticCLIP$_{1}$&\checkmark&\checkmark&\checkmark&\checkmark&0.736&0.754&0.762&0.672&0.831&0.823\\\tabucline[0.5pt]{-}
AquaticCLIP$_{2}$&\checkmark&\checkmark&\checkmark&$\times$&0.721&0.725&0.667&0.651&0.780&0.783\\\tabucline[0.5pt]{-}
AquaticCLIP$_{3}$&\checkmark&\checkmark&$\times$&\checkmark&0.718&0.716&0.731&0.696&0.807&0.811\\\tabucline[0.5pt]{-}
Finetune CLIP&Finetune&Finetune&$\times$&$\times$&0.772&0.847&0.771&0.750&0.853&0.831\\\tabucline[0.5pt]{-}
AquaticCLIP$_{4}$&Finetune&Finetune&\checkmark&$\times$&0.838&\underline{0.890}&\underline{0.821}&0.753&0.892&0.910\\\tabucline[0.5pt]{-}
AquaticCLIP$_{5}$&Finetune&Finetune&$\times$&\checkmark&\underline{0.842}&0.876&0.808&\underline{0.766}&\underline{0.912}&\underline{0.932}\\\tabucline[0.5pt]{-}
AquaticCLIP&Finetune&Finetune&\checkmark&\checkmark&\textbf{0.871}&\textbf{0.923}&\textbf{0.842}&\textbf{0.801}&\textbf{0.934}&\textbf{0.953}\\\tabucline[0.5pt]{-}
\end{tabu}
}}
\end{center}
\label{table1}
%\vspace{-5mm}
\end{table*}
%\vspace{-2mm}
\subsection{Training and Implementation Details}
%\textbf{Instance Detection:} We fine-tuned the existing object detector, RegionCLIP \cite{zhong2022regionclip}, using the CLIP text encoder and ResNet50 as the vision backbone. 
%The fine-tuning was done on publicly available marine datasets containing 228,363 images and 664,411 instances (details are provided in the supplementary material). This fine-tuned model, named %Marine RegionCLIP (MRegionCLIP), followed the same experimental protocols as MarineDET \cite{haixin2023marinedet} and MaineInst \cite{ziqiang2024marineinst}.
%The learning rate was initially set to $5 \times 10^{-2}$, gradually reduced to $5 \times 10^{-4}$ and $5 \times 10^{-6}$ as needed. 
%The fine-tuning was conducted over 9,000 iterations with a batch size of 256 on four A100 GPUs.
%The quantitative object detection results of MRegionCLIP are included in Table \ref{table11}.
%MRegionCLIP was then used to detect instances in our 2M images dataset in a zero-shot fashion.
%The instances detected from each image were then input to the MarineGPT \cite{zheng2023marinegpt} for generating instance-level captions.
The architecture of AquaticCLIP consists of a frozen domain-specific captions generator MarineGPT \cite{zheng2023marinegpt}, the CLIP \cite{radford2021learning} image encoder using the ViT-B/16-224 \cite{dosovitskiy2020image}, and a transformer-based text encoder \cite{radford2019language}. 
We fine-tuned four components: the image encoder, text encoder, prompt-guided vision encoder, and vision-guided text encoder, all using the cross-modal contrastive loss described in Sec. \ref{loss}.
We employed the Adam optimizer \cite{loshchilov2017decoupled} with an initial learning rate of $1 \times 10^{-4}$ and a weight decay of $1 \times 10^{-5}$.
The model was trained for 80 epochs on four A100 GPUs with a batch size of 512. 
We set the number of prompts to 20. 
For fair comparisons, we utilized the same evaluation metrics as used by the SOTA methods.
For the classification tasks, we reported both accuracy and $F_{1}$ measure scores. 
For object detection, we reported mAP$_{50}$ metric.
\textit{Additional details are provided in supplementary material.}
%\vspace{-2mm}
%For pre-training, all images were pre-processed to a size of $512 \times 512$.
%Larger images were resized to 512 on the short side and center-cropped. 
%Data augmentation, including horizontal and vertical flips, was applied to both images and captions. 
%A linear projection head was used to map the text and image embeddings into a 512-dimensional latent space for alignment. 
%The image and text representations were aligned using the cross-modal contrastive loss (Sec. \ref{loss}), and the model was implemented using the PyTorch library.
%In a second experiment, we only fine-tuned our image and text encoders $\Phi_{v}$ and $\Phi_{t}$ using similar settings to CLIP \cite{radford2021learning} with our 2M images and generated text dataset. 
%The settings included a batch size, weight decay of 0.2, a temperature of 0.07, a peak learning rate of $1 \times 10^{-4}$, AdamW optimizer with an initial learning rate of $5 \times 10^{-6}$, and %a cosine decay scheduler.

\subsection{Underwater Datasets and Tasks}
For the zero-shot marine species classification task, we utilized two datasets: Marine Animal Images (MAI) \cite{marine_animal} and Sea Animals Images (SAI) \cite{sea_animal}.
For zero-shot fine-grained classification, we employed three datasets: FishNet \cite{khan2023fishnet}, FishNet Open Images (FNOI) \cite{kay2021fishnet}, and Large-Scale Fish (LSF) \cite{ulucan2020large}.
For zero-shot fine-grained coral species classification, we used the Coral Species Classification (CSC) \cite{coral-species-classification_dataset} and Coral Classification (CC) \cite{coral_classification} datasets.
For object detection and classification, we used four datasets including FishNet, DeepFish \cite{saleh2020realistic}, Brackish \cite{Pedersen_2019_CVPR_Workshops}, and URPC \cite{urpc}.
\textit{Additional details are provided in supplementary material.}
%\vspace{-2mm}
\begin{comment}
\subsection{Downstream Datasets}
For supervised salient object segmentation of underwater images, we used the USOD10K dataset \cite{usod10k}.
For supervised instance segmentation of marine images, we employed the Underwater Image Instance Segmentation (UIIS) \cite{Lian_2023_ICCV}, and for semantic segmentation, we used the SUIM \cite{islam2020suim} dataset. 
For fine-tuned underwater object detection and classification tasks, we used four datasets: FishNet \cite{khan2023fishnet}, DeepFish \cite{saleh2020realistic}, URPC \cite{urpc}, and Brackish \cite{Pedersen_2019_CVPR_Workshops}. 
Lastly, for the biodiversity-related marine object counting task, we used the IOCFish5K \cite{sun2023indiscernible} dataset.
\textbf{\textit{See more details in our supplementary material.}}
\end{comment}

\begin{table*}[t!]
\caption{\textbf{Ablation study} showing the effect of the Textual Descriptions Cleaning Module (TDCM) and the use of instance-level and image-level textual descriptions on AquaticCLIP zero-shot classification performance ($F_{1}$ scores). 
The full AquaticCLIP model, with TDCM and both text levels, delivers the best results across all datasets, especially on MAI (0.871), SAI (0.923), and CC (0.953). 
Variants with components removed or altered exhibit reduced performance, highlighting the importance of each component.}
%\vspace{-7mm}
\begin{center}
\makebox[\linewidth]{
\scalebox{0.90}{
\begin{tabu}{|[2pt]c|c|c|c|[2pt]c|c|c|c|c|c|[2pt]}
\tabucline[0.5pt]{-}
Variants&TDCM&Instance Text&Image Text&MAI&SAI&FishNet&FNOI&LSF&CC\\\tabucline[0.5pt]{-}
AquaticCLIP&\checkmark&\checkmark&\checkmark&\textbf{0.871}&\textbf{0.923}&\underline{0.842}&\textbf{0.801}&\textbf{0.934}&\textbf{0.953}\\\tabucline[0.5pt]{-}
AquaticCLIP$_{6}$&$\times$&\checkmark&\checkmark&\underline{0.854}&0.891&0.804&\underline{0.786}&0.897&\underline{0.934}\\\tabucline[0.5pt]{-}
AquaticCLIP$_{7}$&\checkmark&$\times$&\checkmark&0.853&0.906&0.804&0.765&0.911&0.933\\\tabucline[0.5pt]{-}
AquaticCLIP$_{8}$&\checkmark&\checkmark&$\times$&0.840&0.892&0.823&0.784&0.921&0.915\\\tabucline[0.5pt]{-}
%AquaticVision&$\times$&$\times$&$\times$&0.852&\underline{0.915}&\textbf{0.856}&\underline{0.786}&\underline{0.924}&0.928\\\tabucline[0.5pt]{-}
\end{tabu}
}}
\end{center}
\label{table2}
%\vspace{-8mm}
\end{table*}

\subsection{Ablation Studies}
We conducted ablation studies to highlight the contributions of each component of the proposed AquaticCLIP model. 
The evaluations were performed on zero-shot classification tasks, reporting $F_{1}$ scores across six independent external datasets (MAI, SAI, FishNet, FNOI, LSF, and CC) that were not used during the pre-training phase. \noindent \textbf{1. FrozenCLIP vs. FinetuneCLIP (Table \ref{table1})}: In FrozenCLIP, pre-trained image and text encoders were used to generate visual and text embeddings. 
In FinetuneCLIP, these encoders were fine-tuned using the contrastive pre-training loss on our 2M dataset. 
We observed that fine-tuning resulted in improved performance across all datasets.
\noindent \textbf{2. AquaticCLIP$_{1}$ vs. AcquaticCLIP (Table \ref{table1})}: AquaticCLIP$_{1}$ used frozen image and text encoders, with only the PGVE (Prompt-Guided Vision Encoder) and VGTE (Vision-Guided Text Encoder) fine-tuned. 
Significant performance improvements were observed when all four encoders were fine-tuned in AquaticCLIP. \noindent \textbf{3. Importance of PGVE and VGTE Components (Table \ref{table1})}: In AquaticCLIP$_{2}$ and AquaticCLIP$_{4}$, VGTE was removed, and original textual embeddings were used, while in AquaticCLIP$_{3}$ and AquaticCLIP$_{5}$, PGVE was removed and image-level embeddings from MLP were used. 
Both cases showed performance reductions compared to AquaticCLIP$_{1}$ and AquaticCLIP, highlighting the importance of these components.\\
\noindent \textbf{4. Importance of Textual Description Cleaning Module (TDCM) (Table \ref{table2})}: In AquaticCLIP$_{6}$, removing the TDCM module led to decreased performance compared to the full AquaticCLIP model. \noindent \textbf{5. Importance of Instance-level Text and Image-level Text (Table \ref{table2})}: In AquaticCLIP$_{7}$ and AquaticCLIP$_{8}$, either instance-level or image-level text descriptions were removed. 
Performance dropped in both cases compared to the proposed AquaticCLIP, which utilized both levels of text. 
For fine-grained classification (FishNet, FNOI, LSF), instance-level descriptions performed better, while for coarse-grained classification (MAI, SAI), image-level captions were more effective.
\textit{More ablations are provided in supplementary material.}
%\vspace{-2mm}

\begin{table*}[t!]
\centering
\caption{Zero-shot and supervised classification performance comparison in terms of accuracy and $F_{1}$ score of AquaticCLIP against SOTA VLMs and vision-only models. 
AquaticCLIP consistently outperforms all models across multiple datasets, excelling in both zero-shot and supervised tasks. 
It achieves top $F_{1}$ scores, particularly in MAI, SAI, FishNet, and CC datasets, demonstrating superior generalization and classification accuracy compared to traditional vision-only models and other VLMs. 
Family classification performance is reported for FishNet, while CSC is excluded from supervised methods due to its small size.}
%\vspace{-3mm}
\makebox[\linewidth]{
\scalebox{0.70}{
\begin{tabular}{|c|c|c|c|c|c|c|c|}
\hline
\textbf{Zero-Shot VLMs} &MAI \cite{marine_animal}&SAI \cite{sea_animal}&FishNet \cite{khan2023fishnet}&FNOI \cite{kay2021fishnet}&LSF \cite{ulucan2020large}&CSC \cite{coral-species-classification_dataset}&CC \cite{coral_classification} \\
\hline
Frozen CLIP \cite{radford2021learning}&0.702$|$0.692&0.711$|$0.702&0.663$|$0.651&0.642$|$0.622&0.770$|$0.772&0.809$|$0.783&0.763$|$0.752\\
Finetune CLIP&0.802$|$0.772&0.856$|$0.847&0.770$|$0.771&0.763$|$0.750&0.864$|$0.853&0.862$|$0.841&0.845$|$0.831\\
CoOp \cite{zhou2022learning}&0.853$|$0.822&0.866$|$0.853&0.752$|$0.744&0.764$|$0.752&0.863$|$0.854&0.903$|$0.888&0.868$|$0.866\\
MAPLE \cite{khattak2023maple}&0.861$|$0.834&0.867$|$0.860&0.750$|$0.748&0.774$|$0.769&0.864$|$0.859&0.903$|$0.893&0.881$|$0.876\\
GPT4V \cite{yang2023dawn}&0.831$|$0.811&0.832$|$0.834&0.801$|$0.791&0.758$|$0.743&0.892$|$0.881&0.854$|$0.841&0.881$|$0.876\\
BLIP2 \cite{li2023blip}&0.813$|$0.801&0.821$|$0.818&0.783$|$0.788&0.728$|$0.727&0.893$|$0.880&0.793$|$0.782&0.862$|$0.853\\
MarineGPT \cite{zheng2023marinegpt}&0.862$|$0.844&0.892$|$0.883&\underline{0.823}$|$0.815&0.776$|$0.769&0.912$|$0.905&0.918$|$0.903&0.881$|$0.876\\
AquaticCLIP$_{1}$&0.766$|$0.736&0.746$|$0.754&0.772$|$0.762&0.684$|$0.672&0.844$|$0.831&0.897$|$0.882&0.835$|$0.823\\
AquaticCLIP$_{7}$&\underline{0.879}$|$0.853&\underline{0.912}$|$0.906&0.814$|$0.804&0.773$|$0.765&0.917$|$0.911&\underline{0.944}$|$0.932&\underline{0.938}$|$0.933\\
AquaticCLIP$_{8}$&0.855$|$0.840&0.898$|$0.892&0.821$|$0.823&\underline{0.796}$|$0.784&\underline{0.932}$|$0.921&0.942$|$0.938&0.926$|$0.915\\
AquaticCLIP&\textbf{0.892}$|$\textbf{0.871}&\textbf{0.935}$|$\textbf{0.923}&\textbf{0.850}$|$\textbf{0.842}&\textbf{0.822}$|$\textbf{0.801}&\textbf{0.942}$|$\textbf{0.934}&\textbf{0.968}$|$\textbf{0.964}&\textbf{0.961}$|$\textbf{0.953}\\
\hline 
\textbf{Supervised Vision Models} &MAI \cite{marine_animal}&SAI \cite{sea_animal}&FishNet \cite{khan2023fishnet}&FNOI \cite{kay2021fishnet}&LSF \cite{ulucan2020large}&CSC \cite{coral-species-classification_dataset}&CC \cite{coral_classification} \\
\hline 
ResNet-34 \cite{he2016deep}&0.821$|$0.802&0.731$|$0.726&0.408$|$0.423&0.475$|$0.456&0.761$|$0.756&-&0.833$|$0.821\\
ResNet-50 \cite{he2016deep}&0.830$|$0.811&0.739$|$0.728&0.403$|$0.428&0.504$|$0.488&0.773$|$0.764&-&0.847$|$0.840\\
ResNet-101 \cite{he2016deep}&0.842$|$0.816&0.745$|$0.740&0.363$|$0.354&0.511$|$0.496&0.809$|$0.792&-&0.872$|$0.861\\
ViT-S \cite{alexey2020image}&0.862$|$0.856&0.783$|$0.774&0.379$|$0.367&0.652$|$0.633&0.844$|$0.833&-&0.891$|$0.883\\
ViT-B \cite{alexey2020image}&0.880$|$0.873&0.827$|$0.812&0.429$|$0.412&0.671$|$0.665&0.881$|$0.876&-&0.910$|$0.903\\
ViT-L \cite{alexey2020image}&0.893$|$0.881&0.856$|$0.833&0.484$|$0.476&0.714$|$0.706&0.914$|$0.902&-&0.919$|$0.920\\
BeiT \cite{bao2021beit}&0.881$|$0.873&0.856$|$0.845&0.542$|$0.522&0.704$|$0.688&0.871$|$0.867&-&0.895$|$0.883\\
ConvNeXt \cite{liu2022convnet}&0.847$|$0.852&0.812$|$0.803&0.606$|$0.587&0.714$|$0.702&0.827$|$0.807&-&0.914$|$0.902\\
ConvNeXt \cite{liu2022convnet}+ FL \cite{lin2017focal}&0.881$|$0.870&0.834$|$0.822&0.551$|$0.544&0.738$|$0.722&0.831$|$0.842&-&0.905$|$0.903\\
ConvNeXt \cite{liu2022convnet}+ CB \cite{cui2019class}&0.883$|$0.872&0.851$|$0.842&0.613$|$0.605&0.745$|$0.731&0.870$|$0.855&-&0.917$|$0.905\\
AquaticVision (Linear Probing)&\underline{0.912}$|$\underline{0.890}&\underline{0.945}$|$\underline{0.924}&\underline{0.922}$|$\underline{0.902}&\underline{0.846}$|$\underline{0.833}&\underline{0.942}$|$\underline{0.934}&-&\underline{0.947}$|$\underline{0.931}\\
AquaticCLIP (Linear Probing)&\textbf{0.915}$|$\textbf{0.893}&\textbf{0.951}$|$\textbf{0.944}&\textbf{0.934}$|$\textbf{0.923}&\textbf{0.882}$|$\textbf{0.867}&\textbf{0.968}$|$\textbf{0.961}&-&\textbf{0.963}$|$\textbf{0.958}\\
\hline  
      \end{tabular}
      }}
    \label{table6}
    %\vspace{-5mm}
\end{table*}

\subsection{SOTA Methods for Comparison}
We compared our AquaticCLIP model to a wide range of SOTA methods across various underwater image analysis tasks. 
For zero-shot classification, we compared the performance of our AquaticCLIP model with existing VLMs, including Frozen CLIP \cite{radford2021learning}, Finetune CLIP \cite{radford2021learning}, and prompt-based VLMs like CoOp \cite{zhou2022learning} and MAPLE \cite{khattak2023maple}, as well as GPT4V \cite{yang2023dawn}, BLIP2 \cite{li2023blip}, and MarineGPT \cite{zheng2023marinegpt}.
We fine-tuned CoOp and MAPLE using the original author's provided source codes. 
For supervised classification tasks, we compared AquaticCLIP with models such as ResNet-34/50/101 \cite{he2016deep}, ViT-S/B/L \cite{alexey2020image}, BeiT \cite{bao2021beit}, ConvNeXt \cite{liu2022convnet}, ConvNeXt \cite{liu2022convnet} + Focal Loss (FL) \cite{lin2017focal}, and ConvNeXt \cite{liu2022convnet} + Class-Balanced (CB) \cite{cui2019class}.
To ensure a fair comparison, we used the same settings as FishNet \cite{khan2023fishnet}. 
For object detection, we compared with FasterRCNN \cite{ren2016faster}, YOLOF \cite{chen2021you}, TOOD \cite{feng2021tood}, MarintInst \cite{ziqiang2024marineinst}, MarineDet \cite{haixin2023marinedet}.
\textit{Further details are given in the supplementary material.}
%\vspace{-2mm}
\subsection{Zero-shot Comparisons}
Table \ref{table6} presents the zero-shot classification results and comparisons with SOTA VLM-based methods across seven datasets. 
AquaticCLIP consistently outperformed existing SOTA VLMs by a significant margin on all datasets. 
Significantly, on the CSC dataset, AquaticCLIP achieved a zero-shot performance of 96.80$\%$ accuracy and 96.40$\%$ $F_{1}$ score, the highest across all datasets. 
In more challenging fine-grained fish classification tasks on the FishNet, FNOI, and LSF datasets, AquaticCLIP achieved $F_{1}$ scores of 84.20$\%$, 80.10$\%$, and 93.40$\%$, respectively, due to the incorporation of instance-level captions in the model.
%\vspace{-2mm}
\subsection{Linear Probe Evaluations}
In this experiment, we conducted a linear probe evaluation of the AquaticCLIP model and compared it against several SOTA supervised vision-only models.
For this purpose, the vision encoder of the AquaticCLIP model is kept frozen and only a linear classifier is trained on top of the pre-extracted visual representations.
We also pre-trained our vision-only model, AquaticVision, using contrastive loss on our 2M images dataset in a self-supervised learning setting. 
For this purpose, DINOv2 pre-training paradigm is utilized based on ViT architecture \cite{oquab2023dinov2}.

The supervised comparisons between AquaticCLIP and existing SOTA vision-only models are shown in Table \ref{table6}. 
AquaticCLIP outperformed SOTA vision-based methods, achieving $F_{1}$ scores of 92.30$\%$, 86.70$\%$, and 96.10$\%$ on fine-grained classification datasets. 
We observed that AquaticCLIP also outperforms the AquaticVision model. 
While AquaticVision ranked as the second-best performer, it significantly outperformed other vision-only models due to its contrastive pre-training in a self-supervised learning manner on the 2M aquatic images.

\subsection{Object Detection and Classification Results}
In this experiment, we replaced ResNet-50 in MRegionCLIP with our pre-trained image encoder $\Phi_{v}$ and applied the same fine-tuning settings as discussed in the supplementary. 
We named this model AquaticDet and compared it against SOTA methods.
Table \ref{table11} shows object detection and classification results across four different datasets, with AquaticDet achieving the best mAP$_{50}$ scores across all compared datasets by a significant margin. 
This superior performance is attributed to pre-training on the 2M image-text paired dataset, which allowed AquaticDet to extract highly efficient and effective visual features. The inclusion of the prompt-guided vision encoder and vision-guided text encoder, along with comprehensive captions at both the image and instance levels, contributed to substantial performance improvements, even under challenging conditions.
%\vspace{-2mm}

%\subsection{Object Counting in Underwater Scenes}
%\textbf{\textit{Please see our supplementary material.}}
%For object counting in underwater scenes, we utilized a crowd localization transformer method \cite{cltr_liang2022end}, which includes a CNN-based backbone, a %transformer encoder, a transformer decoder, and a nearest neighbors matching component. 
%In our experiments, we replaced the original backbone and encoder with the AquaticCLIP pre-trained vision encoder $\Psi_{v}$, specifically fine-tuned for aquatic %environments. 
%The rest of the settings followed the original implementation. 
%We refer to our object counting model as AquaticOC.

%Table \ref{table12} presents the object counting results on the IOCFish5K \cite{sun2023indiscernible} dataset and compares AquaticOC with existing SOTA methods. 
%AquaticOC achieved the best results in terms of MAE and MSE, showing a significant improvement over the baseline CLTR model. 
%AquaticOC represents a strong contribution to efficient and accurate object counting in aquatic scenes.

\subsection{Computational Complexity}
We also evaluated the computational complexity of AquaticCLIP during the inference stage across various tasks, including zero-shot classification, supervised object detection and classification, supervised segmentation, and object counting. 
For zero-shot classification, AquaticCLIP took 0.80 seconds, AquaticSAM took 1.23 seconds, AquaticDet took 1.19 seconds, while the AquaticOC model took 1.31 seconds to count objects.
We used the same hardware settings as discussed above.

\begin{table}[t!]
\caption{Object detection and classification results (mAP$_{50}$).}
%\vspace{-7mm}
\begin{center}
\makebox[\linewidth]{
\scalebox{0.78}{
\begin{tabu}{|c|c|c|c|c|}
\tabucline[1.0pt]{-}
Methods&FishNet&DeepFish&Brackish&URPC\\\tabucline[0.5pt]{-}
FasterRCNN \cite{ren2016faster}&0.284&0.814&0.788&0.475\\
YOLOF \cite{chen2021you}&0.672&0.806&0.813&0.511\\
TOOD \cite{feng2021tood}&0.811&0.766&0.805&0.507\\
MRegionCLIP&\underline{0.867}&0.855&\underline{0.842}&0.758\\
MarineInst \cite{ziqiang2024marineinst}&\underline{0.868}&0.854&0.841&\underline{0.779}\\
MarineDet \cite{haixin2023marinedet}&-&-&-&0.706\\
AquaticDet &\textbf{0.903}&\textbf{0.891}&\textbf{0.877}&\textbf{0.837}\\\tabucline[0.5pt]{-}
%DeepLabv3 \cite{deeplab}&0.791&0.812\\
%SUIM-Net \cite{islam2020suim}&0.841&0.869\\
%Mask2Former \cite{cheng2021mask2former}&\underline{0.855}&\underline{0.896}\\
%AquaticSeg&\textbf{0.881}&\textbf{0.921}\\\tabucline[0.5pt]{-}
\end{tabu}
}}
\end{center}
\label{table11}
%\vspace{-10mm}
\end{table}

\section{Conclusion}
\label{sec:conclusion}
\vspace{-1mm}
Current aquatic and underwater VLMs rely on paired image-text data for pre-training.
In this work, we introduced AquaticCLIP, pre-trained using real aquatic image-text pairs and additional generated textual descriptions at both image and instance levels. 
To achieve this, we built a 2M image-text paired dataset sourced from various online repositories. 
We proposed a novel vision-language alignment model where the vision encoder is guided by learned prompts, and the text encoder benefits from visual prompts. 
Both lightweight encoders are pre-trained using cross-modal contrastive supervision for enhanced vision-language alignment.
AquaticCLIP was evaluated across a diverse set of marine vision tasks, including zero-shot fine-grained object classification, fine-tuned instance and semantic segmentation, and object detection, and counting.
Our model consistently delivered superior results compared to existing SOTA methods designed specifically for marine environments, demonstrating its robustness and effectiveness across multiple aquatic vision tasks.

\ifCLASSOPTIONcaptionsoff
  \newpage
\fi

\bibliographystyle{IEEEtranS}
\bibliography{main}

\ifCLASSOPTIONcaptionsoff
  \newpage
\fi

\vspace{-15mm}
\begin{IEEEbiography}
	[{\includegraphics[width=1in,height=2.0in,clip,keepaspectratio]{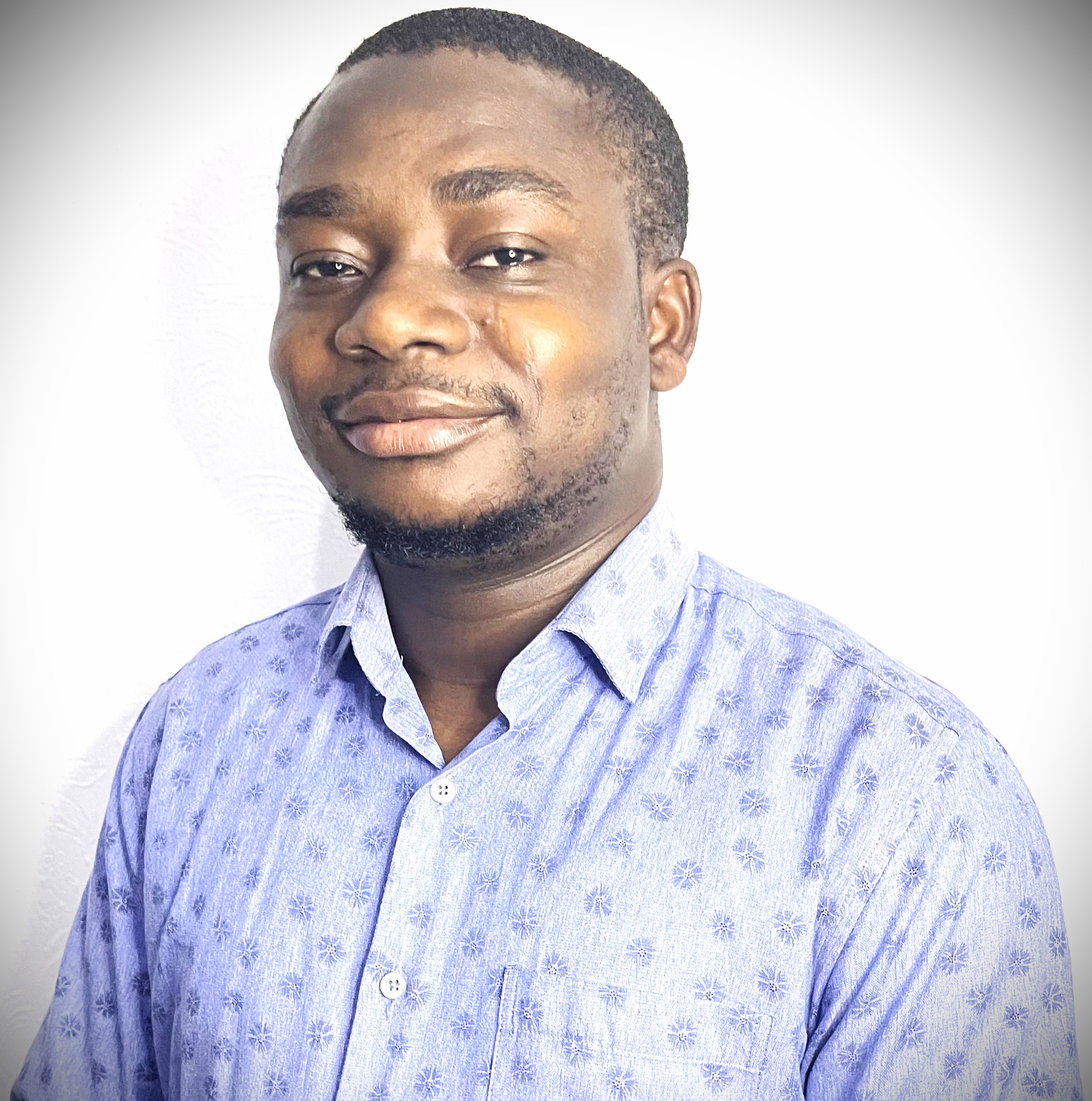}}]{Basit Alawode}
received his BSc degree in electrical engineering from the Obafemi Awolowo University, Nigeria, in 2013. 
He further obtained his MSc degree in electrical engineering at the King Fahd University of Petroleum and Minerals, Saudi Arabia in 2020. 
He is currently pursuing his Ph.D. degree in computer science and engineering degree at Khalifa University of Science and Technology, UAE.
His research interests include visual object tracking and computer vision.
\end{IEEEbiography}
\vspace{-15mm}
\begin{IEEEbiography}
	[{\includegraphics[height=2.2in, width=1.0in,clip,keepaspectratio]{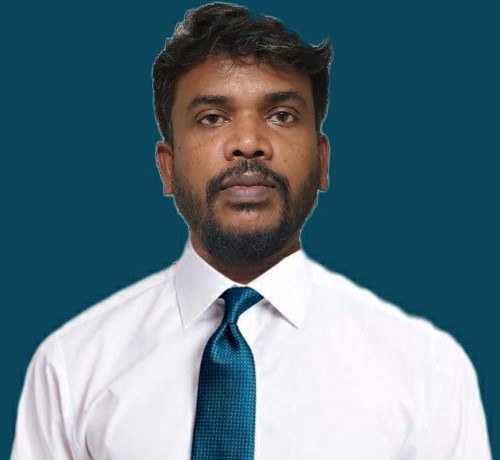}}]
{Iyyakutti Iyappan Ganapathi} is currently a postdoctoral fellow at Department of Electrical Engineering and Computer Science, Khalifa University, Abu Dhabi, UAE. He previously worked as an Assistant Professor at Woosong University in South Korea. He earned his PhD degree from the Indian Institute of Technology Indore, India. 3D image processing, biometrics, computer vision, and machine learning are among his research interests.
\end{IEEEbiography}
\vspace{-15mm}
\begin{IEEEbiography}
	[{\includegraphics[width=1in,height=2.0in,clip,keepaspectratio]{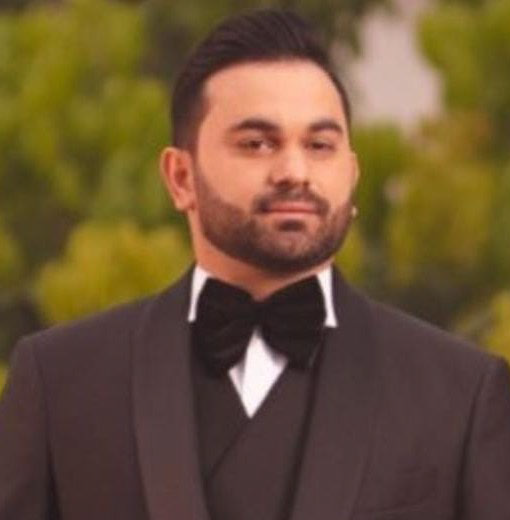}}]{Sajid Javed}
is a faculty member at Khalifa University (KU), UAE. 
Prior to that, he was a research fellow at KU from 2019 to 2021 and at the University of Warwick, UK, from 2017-2018. 
He received his B.Sc. degree in computer science from the University of Hertfordshire, UK, in 2010. 
He completed his combined Master’s and Ph.D. degrees in computer science from Kyungpook National University, Republic of Korea, in 2017.
\end{IEEEbiography}
\vspace{-15mm}
\begin{IEEEbiography}
[{\includegraphics[width=1in,height=1.35in,clip,keepaspectratio]{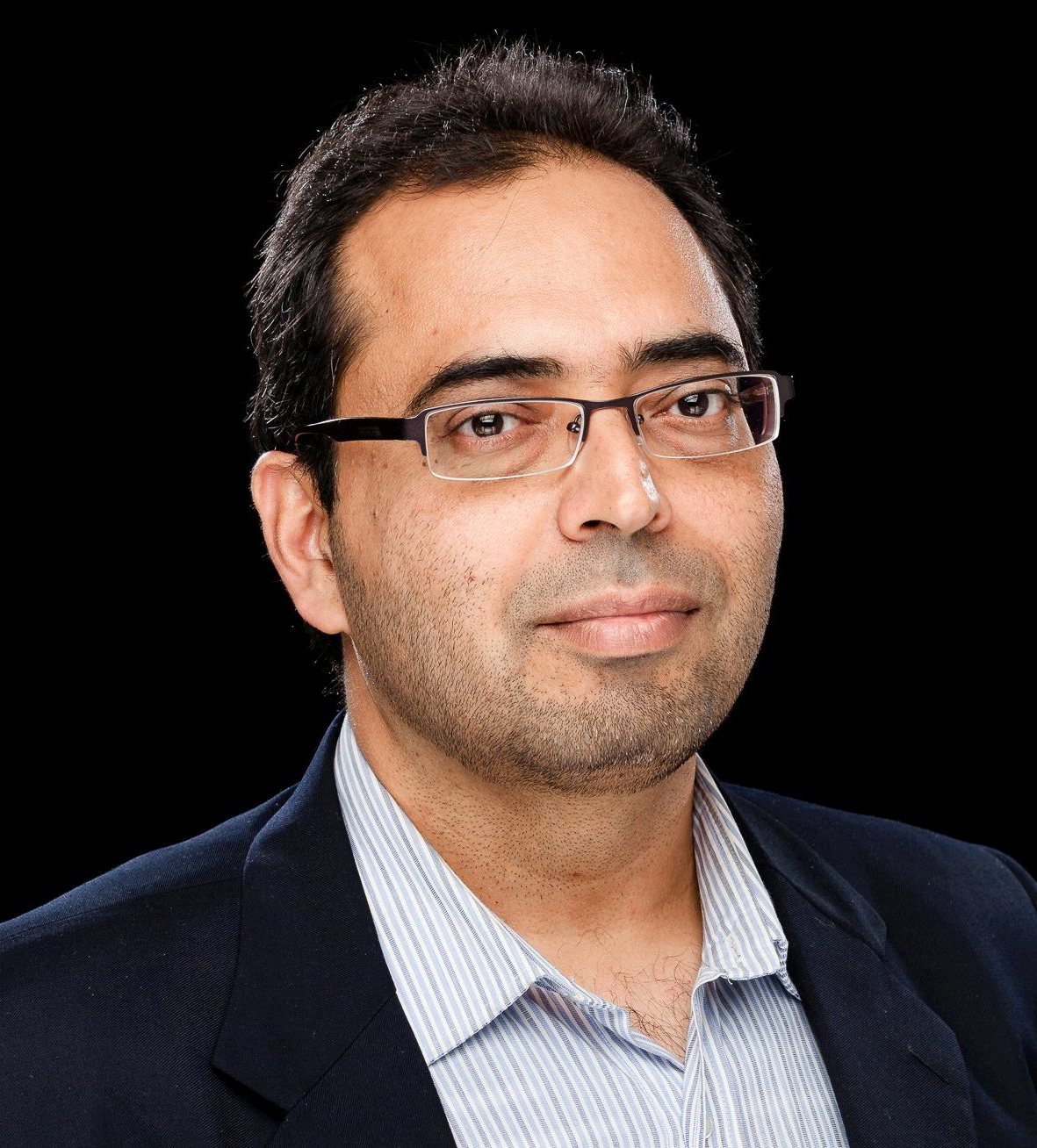}}]{Arif Mahmood}
is currently a professor at the Department of Computer Science at ITU, Pakistan, and the director of the computer vision lab.
He received his M.Sc. and Ph.D. degree in computer science from LUMS, Pakistan, in 2003 and 2011.
He has also worked as a Research Assistant Professor with the School of Mathematics and Statistics (SMS), University of Western Australia (UWA).
His research interests are face recognition, object classification, human-object interaction detection, and abnormal event detection. 
\end{IEEEbiography}
\vspace{-12mm}
\begin{IEEEbiography}
	[{\includegraphics[width=1in,height=1.35in,clip,keepaspectratio]{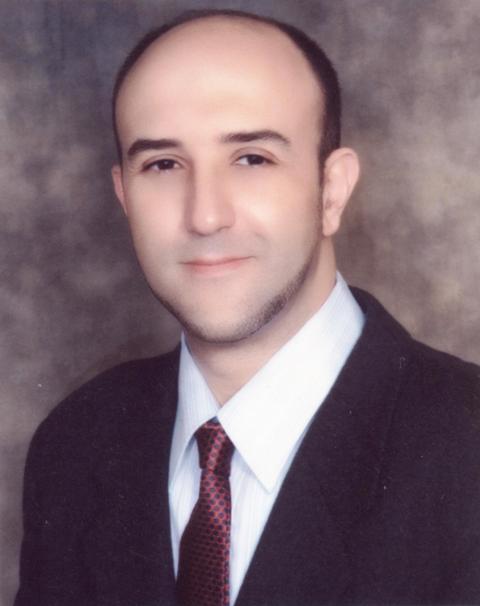}}]{Naoufel Werghi}
is a Professor at the department of computer science in Khalifa University for Science and Technology, UAE.
He received his Habilitation and PhD in Computer Vision from the University of Strasbourg.
His main research area is 2D/3D image analysis and interpretation, where he has been leading several funded projects related to biometrics, medical imaging, remote sensing, and
intelligent systems. 
\end{IEEEbiography}

\newpage
\huge
\textbf{Supplementary Material \\ AquaticCLIP: A Vision-Language Foundation Model for Underwater Scene Analysis}
\normalsize

\section*{Table of Contents}
\begin{enumerate}
    \item  \href{dataset}{Construction, Curation, and Cleaning of a 2M Aquatic Image-Text Paired Dataset (Sec. \ref{dataset}).}
    \item  \href{training}{Additional Training and Implementation Details (Sec. \ref{training}).}
    \item  \href{ablations}{More Ablation Studies (Sec. \ref{ablations}).}
    \item  \href{mrgionclip}{MRegionCLIP Pre-training for Instance Detection (Sec. \ref{mrgionclip}).}
    \item  \href{unsupervised}{Unsupervised Generation of Image and Instance-Level Descriptions (Sec. \ref{unsupervised}).}
    \item  \href{datasets}{Underwater Datasets and Tasks (Sec. \ref{datasets}).}
    \item  \href{sota}{SOTA Methods for Comparison (Sec. \ref{sota}).}
    \item  \href{sotadetails}{SOTA Methods Training Details (Sec. \ref{sotadetails}).}
    \item  \href{metrics}{Evaluation Metrics (Sec. \ref{metrics}).}
    \item  \href{inference}{Zero-shot Inference (Sec. \ref{inference}).}
    \item  \href{cross}{Zero-shot Cross-Modal Retrieval Results (Sec. \ref{cross}).}
    \item  \href{objdetect}{Underwater Object Detection and Classification Results (Sec. \ref{objdetect}).}
    \item  \href{segmentation}{Underwater Scene Segmentation Results (Sec. \ref{segmentation}).}
    \item  \href{counting}{Results of Object Counting in Underwater Scenes (Sec. \ref{counting}).}
    \item  \href{supervised}{Supervised AquaticCLIP Model: Linear Probe Evaluations (Sec. \ref{supervised}).}
    \item  \href{vision}{AquaticVision: Vision-Only Model Pre-training Details (Sec. \ref{vision}).}
    \item  \href{visual}{ Visual Results (Sec. \ref{visual}).}
    \item  \href{discussion}{Why AquaticCLIP Performance is Better? (Sec. \ref{discussion}).}    
\end{enumerate}

\begin{figure*}[t!]
\centering
\includegraphics[width=\linewidth]{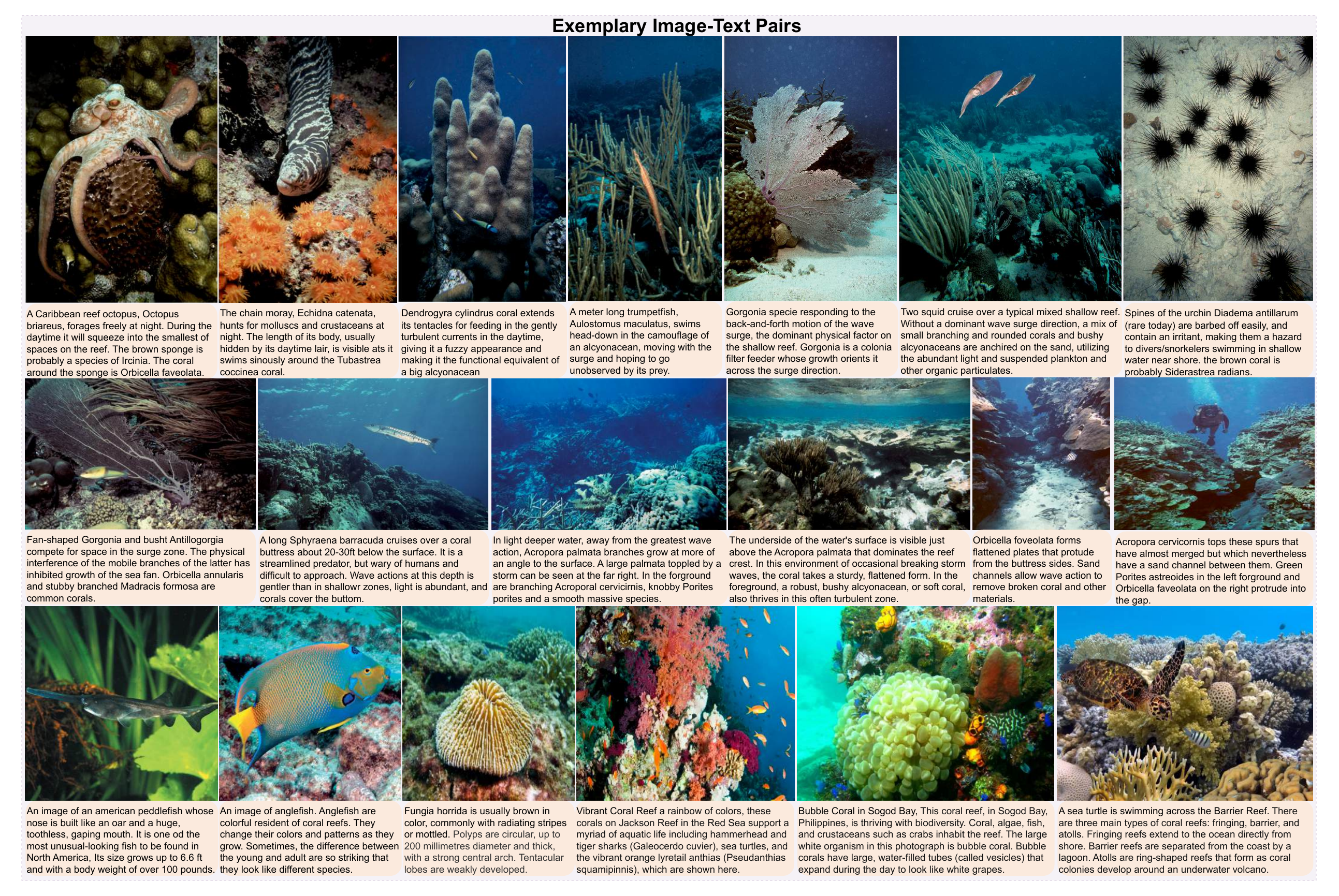}
\caption{Exemplary image-text pairs from our 2 million aquatic image-text paired dataset.}
\label{fig:image_caption_gen1}
\end{figure*}

\begin{figure*}[t!]
\centering
\includegraphics[width=\linewidth]{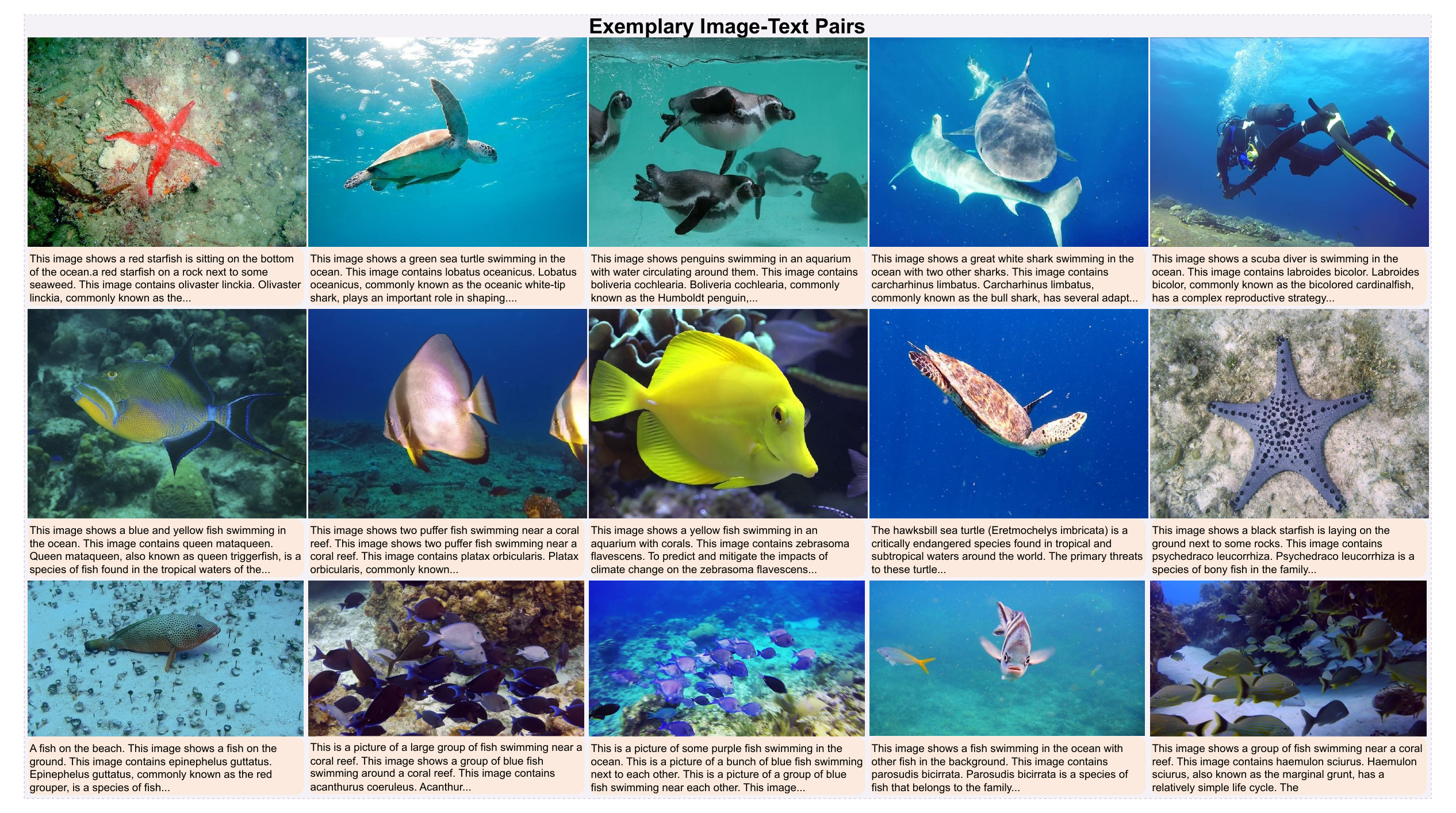}
\caption{Exemplary image-text pairs from our 2 Million aquatic image-text paired dataset.}
\label{fig:image_caption_gen2}
\end{figure*}

\section{Construction, Curation, and Cleaning of a 2M Aquatic Image-Text Paired Dataset}
\label{dataset}
Our aquatic dataset consists of a diverse collection of 2 million image-text pairs collected from various online and heterogeneous sources. 
These include YouTube and Netflix videos, a variety of textbooks on marine biology and oceanography, content on underwater species like fish and sharks, marine articles, as well as repositories like Coral of the World \cite{veron2016corals}, Fishes of Australia \cite{van2014family, schodde1997zoological, merrick2006australasian,shelley2017revision}, Marine Twitter, and National Geographic (NatGeo) \cite{doe2023coral, national_geographic_coral, ng_coral_bleaching, ng_ocean_acidification,ng_great_barrier_reef,ng_coral_conservation,ng_ocean_biodiversity}. 
In total, the dataset contains 20.3 million instances, with at least one instance per image and an average of 10.3 instances per image.
To supplement freely available resources, we allocated \$6550 USD toward subscriptions, PDFs, CDs of marine biology books, and NatGeo magazines. 

\begin{figure*}[t!]
\centering
\includegraphics[width=0.9\linewidth]{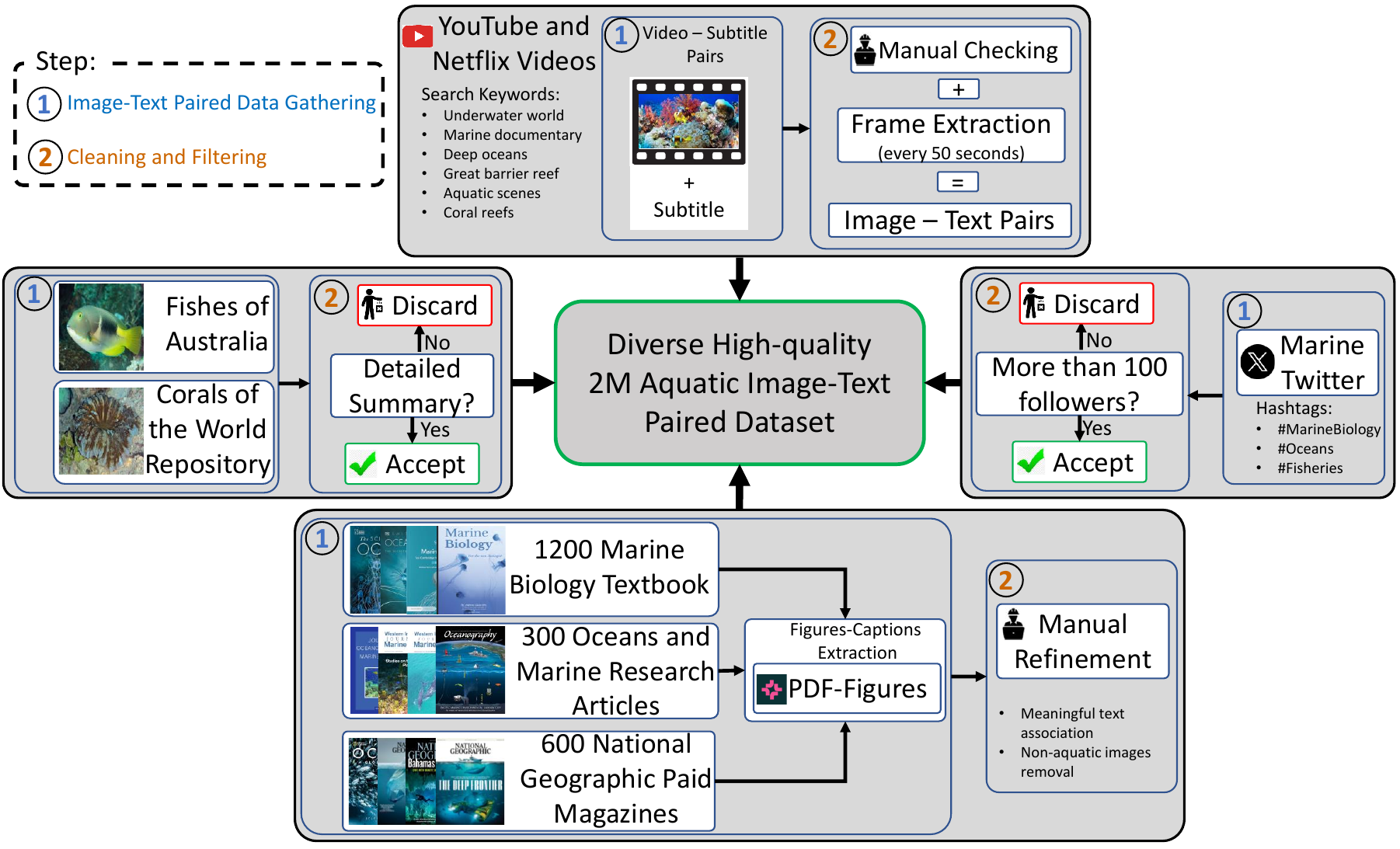}
\caption{Workflow for constructing and cleaning a diverse, high-quality 2 million aquatic image-text paired dataset for Aquatic-CLIP model pre-training. 
This process involves two main steps: (1) Image-Text Paired Data Gathering from sources such as YouTube and Netflix videos, marine biology textbooks, online repositories, and social media, and (2) Cleaning and Filtering through manual checks, frame extraction, caption refinement, and content validation to ensure dataset accuracy and relevance..}
\label{fig:image_caption_gen3}
\end{figure*}

\noindent Figs. \ref{fig:image_caption_gen1}-\ref{fig:image_caption_gen2} show some of the exemplar image-text pairs from our 2M dataset.
Fig. \ref{fig:image_caption_gen3} illustrates the dataset construction pipeline, which includes two key steps: the collection of image-text pairs from multiple sources, followed by meticulous cleaning and filtering to ensure dataset quality.

\subsection{YouTube Videos} 
We reviewed several hundred documentary videos on underwater, marine, and ocean topics, using keywords such as ``\textit{underwater world}'', ``\textit{underwater paradise}'', ``\textit{Marine Life and Ocean Animals}'', ``\textit{Our Planet}'', ``\textit{Wonders of Oceans}'', ``\textit{underwater killers}'',  ``\textit{sea creatures}'', ``\textit{ocean animals for kids}'', ``\textit{kingdoms of marine life}'', ``\textit{Reef Life of the Andaman}'', ``\textit{The Big Blue Ocean stories}'', ``\textit{marine documentary}'', ``\textit{deep oceans}'', ``\textit{great barrier reef}'', ``\textit{aquatic scenes}'',  ``\textit{coral reefs}'', ``\textit{Gorgeous Underwater World of the Philippines}'', ``\textit{Mysterious hunters of the deep sea}'', ``\textit{Dark side of the ocean}'',``\textit{Beautiful Coral Reefs}'', ``\textit{Mysterious of the Twilight Zone}''.\\

\subsection{Netflix Documentaries} 
Using a paid subscription, we explored hundreds of hours of documentary videos, using the following titles: ``\textit{My Octopus Teacher}'', ``\textit{Last Breath}'', ``\textit{Wonders of the Reef}'', ``\textit{Seaspiracy}'', ``\textit{Mission blue}'', ``\textit{The deepest breath}'', ``\textit{Our planet}'', ``\textit{chasing coral}'', ``\textit{Microworlds reef}'', ``\textit{Secret world of sound}'', ``\textit{Night of earth}'', ``\textit{Disney Nature Oceans}'', ``\textit{A Plastic Ocean}'', ``\textit{Our Oceans}''.\\

\noindent Subtitles from these videos were used to create aligned image-text pairs, which were manually reviewed and refined. 
We extracted unique frames every 50 seconds, which presented challenges like low visibility, motion blur, background clutter, and color distortions. 
In total, we captured 0.8 million image-text pairs from YouTube videos and 0.5 million pairs from Netflix documentaries.

\subsection{Textbooks on Marine Biology, Oceanography, Underwater Creatures and Species, Fish, and Sharks} 
We used a total of 1,200 different textbooks on marine biology, ocean science, underwater creatures and species, fish, sea life, and coral reefs, yielding 0.4 million image-text pairs. Below are 50 example books:

\noindent 1. \textit{Exploring the World of Aquatic Life by John Dawes.}
2. \textit{Marine Biology Basics.}
3. \textit{New Species Described in Corals of the World by Veron.}
4. \textit{Introduction to Marine Biology, 3rd Edition by George.}
5. \textit{Marine Biology: A Very Short Introduction by Philip.}
6. \textit{The soul of an octopus: A surprising exploration into the wonder of consciousness by Montgomery.}
7. \textit{The World Beneath: The Life and Times of Unknown Sea Creatures and Coral Reefs by Richard.}
8. \textit{Becoming a Marine Biologist by Morell.}
9. \textit{Handbook of Whales, Dolphins, and Porpoises of the World by Mark.}
10. \textit{Marine Biology: An Ecological Approach by James.}
11. \textit{Citizens of the Sea: Wondrous Creatures from the Census of Marine Life by Nancy.}
12. \textit{Deep: Freediving, Renegade science, and what the ocean tells us about ourselves by Nestor.}
13. \textit{Marine Biology For The Non-Biologist by Andrew.}
14. \textit{Marine Biology: Comparative Ecology of Planet Ocean by Roberto.}
15. \textit{Marine Ecology by Michel.}
16. \textit{Oceanography and marine biology: an introduction to marine science by David.}
17. \textit{Watching giants: the secret lives of whales by Kelsey.}
18. \textit{The ocean of life: The fate of man and the sea by Callum.}
19. \textit{Sink Like Fish: How Sound Rules Life Underwater by Amorina.}
20. \textit{The Marine World by Frances.}
21.	\textit{Marine Biology: Function, Biodiversity, Ecology by Jeffrey S. Levinton.}
22.	\textit{Introduction to the Biology of Marine Life by John Morrissey and James L. Sumich.}
23.	\textit{Marine Biology by Peter H. Raven and George B. Johnson.}
24.	\textit{Biology of the Marine Environment by John D. H. Connell.}
25.	\textit{Marine Biology: A Very Short Introduction by Philip V. Wells.}
26.	\textit{Marine Biology for the Non-Biologist by Anne E. McCarthy.}
27.	\textit{Fundamentals of Marine Biology by Paul F. McCarthy.}
28.	\textit{Marine Conservation Biology by H. A. J. W. Davis.}
29.	\textit{Marine Biotechnology by David L. K. A. V. Hunter.}
30.	\textit{Tropical Marine Ecology by R. G. K. Smiley.}
31.	\textit{Marine Biogeochemistry: A Multidisciplinary Approach by K. J. C. Browne.}
32.	\textit{Fisheries Biology, Assessment and Management by A. A. G. J. K. Smith.}
33.	\textit{Marine Biotechnology: Methods and Protocols by C. B. H. R. S. Farago.}
34.	\textit{Aquatic Ecosystems: Trends and Global Perspectives by A. C. W. F. R. B. Stokes.}
35.	\textit{Marine Ecotoxicology by T. B. R. R. O. G. C. Chikaram.}
36.	\textit{Marine Biogeography: An Overview by M. C. E. D. E. K. P. Chen.}
37.	\textit{Ocean Biogeochemistry by M. R. M. B. D. B. F. N. Andrews.}
38.	\textit{Marine Ecology: A Comprehensive Study of the Ecology of Marine Organisms by M. J. E. S. McMahon.}
39.	\textit{Aquaculture: Principles and Practices by D. S. S. H. L. P. T. J. S. M. G. Thomas.}
40.	\textit{Marine Resource Management: A Global Perspective by L. J. T. S. B. W. Smith.}
41.	\textit{Marine Policy: A Comprehensive Overview by R. B. M. W. G. H. J. K. M. W. D. J. H. Mark.}
42.	\textit{Marine Ecosystem Management by D. J. M. J. K. R. K. M. S. T. J. Stokes.}
43.	\textit{The Diversity of Fishes: Biology, Evolution, and Ecology.}
44.	\textit{Marine Invertebrates of the Pacific Northwest.}
45.	\textit{Coral Reef Fishes: Dynamics and Diversity in a Complex Ecosystem.}
46.	\textit{Introduction to Marine Biology.}
47.	\textit{Fishes of the Gulf of Maine.}
48.	\textit{Marine Mammals: Evolutionary Biology.}
49.	\textit{Marine Algae of the Pacific Northwest.}
50.	\textit{The Ecology of Fishes on Coral Reefs.}

\subsection{Oceans and Marine Research Articles} 
We gathered 178K image-text pairs from 300 marine science articles obtained through Google.

\subsection{National Geographic (NatGeo) Subscribed Magazines} 
We collected 100K image-text pairs from 600 subscribed NatGeo magazines.
Some sample magazine titles include ``\textit{Corals Reef}'', ``\textit{underwater exploration}'', ``\textit{national geographic explorer}'', ``\textit{national geographic sea change}'', ``\textit{Oceanography}'', ``\textit{Adventure}'', ``\textit{Science of the Sea}'', ``\textit{Magazine: The Ocean}'', ``\textit{The Wonders of the Sea}'',``\textit{Oceans and the Environment}'',``\textit{Underwater World}'',``\textit{Marine Conservation}'',``\textit{The Blue Planet}'',``\textit{Exploring the Deep Sea}'',``\textit{Life Underwater}''.

Figures and captions were extracted using PDF-Figures 2.0 tool \cite{clark2016pdffigures}, and we manually refined the data to ensure that the selected images had meaningful text and captions. Non-aquatic images were discarded.

\subsection{Corals of the World and Fishes of Australia} 
These repositories are publicly available. We extracted 15K image-text pairs from these resources and manually verified that the images depicted aquatic scenes and had detailed, meaningful textual descriptions.

\subsection{Marine Twitter} 
We used Twitter to search for relevant content using hashtags such as $\#$MarineBiology, $\#$Oceans, and $\#$Fisheries, considering only channels with over 100 followers. 
We collected 7K image-text pairs from this source after a thorough cleaning and filtering process.

\section{Additional Training and Implementation Details}
\label{training}
The AquaticCLIP model architecture includes a frozen domain-specific caption generator (MarineGPT \cite{zheng2023marinegpt}) to produce unsupervised additional textual descriptions, the CLIP \cite{radford2021learning} image encoder with ViT-B/16-224 \cite{dosovitskiy2020image} for extracting patch-level embeddings, and a transformer-based text encoder with a 76-token maximum sequence length \cite{radford2019language} for textual embeddings. 
We fine-tuned four components: the image encoder, text encoder, prompt-guided vision encoder, and vision-guided text encoder, utilizing cross-modal contrastive loss as described in Section 3.6.
The Adam optimizer \cite{loshchilov2017decoupled} was used with an initial learning rate of $1 \times 10^{-4}$ and weight decay of  $1 \times 10^{-5}$.
The model was trained over 80 epochs on four A100 GPUs, with a batch size of 512. 
We set 20 prompts for the model. 
During pre-training, all images were resized to $512 \times 512$, with larger images resized on the short side to 512 and center-cropped. 
Data augmentation, including horizontal and vertical flips, was applied to both images and captions. 
A linear projection head mapped the text and image embeddings into a 512-dimensional latent space for alignment. 
Image and text representations were aligned using the cross-modal contrastive loss (Section 3.6), and the model was implemented with PyTorch.

In a second experiment, we fine-tuned only the image and text encoders, $\Phi_{v}$ and $\Phi_{t}$, using a setup similar to CLIP \cite{radford2021learning} with our 2M image-text paired dataset. 
This configuration included a batch size, a weight decay of 0.2, a temperature setting of 0.07, a peak learning rate of $1 \times 10^{-4}$, the AdamW optimizer with an initial learning rate of $5 \times 10^{-6}$, and a cosine decay scheduler.

\section{More Ablation Studies}
\label{ablations}
Figs. \ref{fig:frozen_clip}-\ref{fig:aquatic_clip5} show the architectural ablation figures highlighting the importance of the proposed Prompt-guided Vision Encoder (PGVE) and Vision-guided Text Encoder (VGTE).
Please refer to Table \textcolor{cyan}{1} of the main manuscript.

\begin{enumerate}
    \item \textbf{Comparison of Heterogeneous Textual Sources (Table \ref{table1})}: We evaluated AquaticCLIP with text descriptions generated by MarineGPT (MGPT) \cite{zheng2023marinegpt}, GPT4V \cite{yang2023dawn}, and BLIP2 \cite{li2023blip}.
MGPT achieved the best results due to its specific pre-training in the aquatic domain, while GPT4V and BLIP2 were trained on general-domain data. 
Combining ground truth with MGPT-generated texts further improved performance, as the two sources complemented each other in providing more accurate textual descriptions.

\item \textbf{Effect of the Number of Learnable Prompts on Performance (Table \ref{table2})}: We varied the number of learnable prompts $n_{r}$  in the Prompt-Guided Vision Encoder (PGVE) from 5 to 30, as shown in \ref{table2}. 
Performance increased with more prompts, peaking at 0.842 for $n_{r}=20$ on the FishNet dataset. 
Fewer prompts led to incomplete visual feature representation, while more than 20 prompts added redundancy without further performance gains.
\item \textbf{Performance Variation with Different Top-p$\%$ Keywords (Table \ref{table2})}: In the image-text caption cleaning module, we experimented with retaining different percentages of top-matching keywords, ranging from 90$\%$  to 10$\%$ , as shown in Table \ref{table2}. 
Optimal performance occurred when 20$\%$ of the top keywords were retained. 
Higher percentages likely included noisy keywords, reducing performance, while lower percentages discarded too many vital keywords, leading to a performance decrease.
\end{enumerate}

\begin{table}[t!]
\caption{\textbf{Ablation study} comparing ground truth (GT) text with generated text from MarineGPT (MGPT), GPT4V, and BLIP2 for zero-shot classification ($F_{1}$ scores). 
The combination of GT and MGPT yields the best performance across all datasets, with top scores on MAI (0.871), SAI (0.923), and CC (0.953). 
Generated text from GPT4V and BLIP2 shows lower performance, highlighting the advantage of using domain-specific MGPT with GT..}
\begin{center}
\makebox[\linewidth]{
\scalebox{0.88}{
\begin{tabu}{|[2pt]c|c|c|c|c|c|[2pt]}
\tabucline[0.5pt]{-}
Variants&GT&GT+MGPT&MGPT&GPT4V&BLIP2\\\tabucline[0.5pt]{-}
MAI&\underline{0.861}&\textbf{0.871}&0.844&0.811&0.801\\\tabucline[0.5pt]{-}
SAI&\underline{0.902}&\textbf{0.923}&0.883&0.834&0.818\\\tabucline[0.5pt]{-}
FishNet&\underline{0.821}&\textbf{0.842}&0.815&0.791&0.788\\\tabucline[0.5pt]{-}
FNOI&\underline{0.785}&\textbf{0.801}&0.769&0.743&0.727\\\tabucline[0.5pt]{-}
LSF&\underline{0.917}&\textbf{0.934}&0.905&0.881&0.880\\\tabucline[0.5pt]{-}
CC&\underline{0.936}&\textbf{0.953}&0.917&0.876&0.853\\\tabucline[0.5pt]{-}
\end{tabu}
}
}
\end{center}
\label{table1}
\end{table}

\begin{table}[t!]
\caption{\textbf{Ablation study} on AquaticCLIP, detailing the impact of varying the number of learnable prompts $n_{r}$ in the Prompt-Guided Vision Encoder (PGVE) and the percentage of top-p$\%$ keywords. 
$F_{1}$ scores for zero-shot classification on the FishNet dataset indicate optimal performance at 20 prompts and 30$\%$ keyword retention, both achieving an $F_{1}$ score of 0.842. Changes in these parameters lead to minor performance variations.}
\begin{center}
\makebox[\linewidth]{
\scalebox{0.80}{
\begin{tabu}{|c|c|c|c|c|c|c|}
\tabucline[1.5pt]{-}
Prompts ($n_{r}$)&5&10&15&20&25&30\\\tabucline[0.5pt]{-}
AquaticCLIP&0.822&0.835&0.839&\textbf{0.842}&0.840&0.841\\\tabucline[1.5pt]{-}
Top-p$\%$ keywords&90&70&50&30&20&10\\\tabucline[0.5pt]{-}
AquaticCLIP&0.791&0.811&0.830&0.837&\textbf{0.842}&0.826\\\tabucline[1.5pt]{-}
\end{tabu}
}}
\end{center}
\label{table2}
\end{table}

\section{MRegionCLIP Pre-training for Instance Detection}
\label{mrgionclip}
The instance detection method, MRegionCLIP, is pre-trained on publicly available datasets (see Table \ref{t:pretraining_datasets}) and then applied as a frozen instance detector to our 2M aquatic dataset. 

We fine-tuned an existing object detector, RegionCLIP \cite{zhong2022regionclip}, using the frozen CLIP text encoder and ResNet50 as the vision backbone. 
Similar to methods like RegionCLIP \cite{zhong2022regionclip}, MarineDET \cite{haixin2023marinedet}, and MarineInst20M \cite{ziqiang2024marineinst}, the COCO Caption dataset was used for contrastive learning. 
For marine-specific pre-training, we fine-tuned on publicly available marine datasets, which contain 228,363 images and 664,411 instances.
A powerful region proposal network (RPN) and a frozen image encoder from RegionCLIP were used.
This fine-tuned model, named Marine RegionCLIP (MRegionCLIP), followed the same experimental protocols as MarineDET \cite{haixin2023marinedet} and MarineInst20M \cite{ziqiang2024marineinst}.
The initial learning rate was set to $5 \times 10^{-2}$, then gradually reduced to $5 \times 10^{-4}$ and $5 \times 10^{-6}$ as needed. 
Fine-tuning was performed over 9,000 iterations with a batch size of 256 on four A100 GPUs.
Quantitative object detection results for MRegionCLIP are provided in Table 10 of the main paper. 
MRegionCLIP was then utilized in zero-shot settings to detect instances in our 2M image-text paired dataset, and these detected instances were used for instance-level caption generation by MarineGPT \cite{zheng2023marinegpt}.

\begin{table*}[t!]
    \centering
   \caption{Publicly available underwater datasets used for pre-training MRegionCLIP in instance detection. Each dataset includes details on diversity, original task, number of images, instances, and average instances per image. The combined dataset totals 228,363 images and 664,678 instances, averaging 2.91 instances per image.}
%\twemoji{right arrow} denotes the instance masks are \textbf{generated by models} based on various prompts or automatically (no prompts). \twemoji{man office worker} denotes instance masks are annotated by \textbf{human annotators} based on our written labeling tool.
   
   \adjustbox{width=0.95\textwidth}{
    \begin{tabular}{ l | c | c | c | c | c }
    \hline
    Dataset & Diversity & Original task and Motivation & Images & Instances & Average \\
    \hline
     Online images \cite{flickr,getty_images,shutterstock} & High & Image collection (human labeled) & 35,175 & 194,010 & 5.52 \\ 
     Aquarium \cite{aquarium_dataset} & Medium & Underwater object detection & 632 & 4,182 & 6.62 \\ 
     FathomNet \cite{fathomnet_Katija2022} & High & Underwater and deep-sea object detection & 69,909 & 121,329 & 1.74 \\ 
     FLOW \cite{flow_Cheng_2021_ICCV} & Medium & Litter detection & 1,825 & 3,850 & 2.39 \\ 
     MarineDET \cite{haixin2023marinedet} & High & Open-marine object detection & 22,679 & 39,243 & 1.73 \\ 
     MarineFouling \cite{marine_fouling} & Low & Biological fouling detection & 221 & 508 & 2.30 \\ 
     OZFish \cite{ozfish} & Meduim & Underwater fish detection & 6,235 & 38,875 & 6.23 \\ 
     TACO \cite{taco} & Medium & Litter detection & 1,109 & 2,656 & 2.39 \\ 
     TrashCAN \cite{trashcan} & Medium & Underwater trash detection & 6,465 & 9,855 & 1.52 \\ 
     UDD/DUO \cite{duo} & Medium & Underwater object detection & 2,170 & 13,090 & 6.03 \\ 
     Underwater\_Garbage \cite{underwater_trash_det} & Medium & Underwater garbage detection & 4,542 & 9,386 & 2.07 \\ 
     EOL \cite{eol} & High & Species identification & 23,141 & 80,128 & 3.46 \\ 
     FishDB \cite{fishdb} & Medium & Fish species identification & 9,905 & 18,914 & 1.91 \\ 
     HK-Reef-Fish \cite{hk_reef_fish} & Low & Fish identification & 729 & 1,985 & 2.72 \\ 
     ImageNet\_Sub \cite{imagenet_sub} & Low & Scene classification & 3,987 & 7,175 & 1.78 \\ 
     Oceanic\_Life \cite{oceanic_life} & High & Collection of marine life imagery & 5,029 & 20,811 & 4.14 \\ 
     Reef-Life-Survey \cite{reef_life_survey} & High & Marine creature identification & 7,075 & 12,502 & 1.77 \\ 
     Reeflex \cite{reeflex} & High & Marine creature identification & 15,088 & 61,656 & 4.09 \\ 
     Sea Animals \cite{sea_animal} & Medium & Sea animal classification & 3,080 & 7,448 & 2.42 \\ 
     WildFish++ \cite{wildfishpp} & High & Fine-grained fish classification & 9,367 & 17,075 & 1.82 \\ 
     \hline
     \textbf{Total} & \textbf{High} & \textbf{Instance detecting pre-training dataset} & \textbf{228,363} & \textbf{664,678} & \textbf{2.91} \\ 
    \hline
    \end{tabular}}
    \label{t:pretraining_datasets}
\end{table*}

\section{Unsupervised Generation of Image and Instance-Level Descriptions}
\label{unsupervised}
Using our pre-trained object detector, MRegionCLIP, we detect objects as instances in each image within our 2M dataset. 
Each detected object instance is input into the MarineGPT model for caption generation. 
Specifically, we used the following prompt template: ``The image is $<\textbf{image}>$. Describe the object in this image:'', where $<\textbf{image}>$ is the
image token.
At the image level, each complete image $\textbf{I}_{i}$ is fed into MarineGPT to generate corresponding textual descriptions.

\begin{figure*}[t!]
\centering
\includegraphics[width=0.9\linewidth]{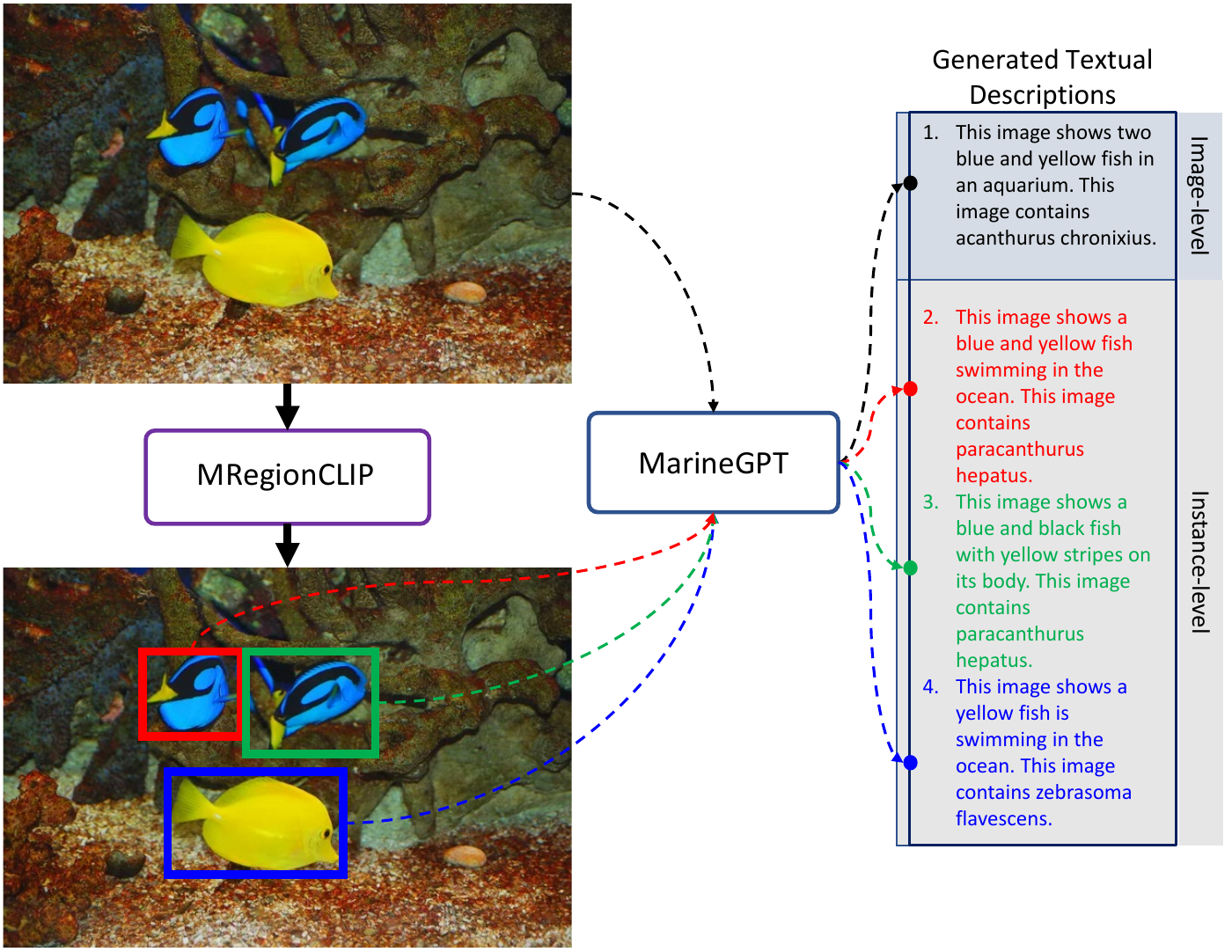}
\caption{Process of unsupervised image-and instance-level caption generation on a sample image using MRegionCLIP for object detection and MarineGPT \cite{zheng2023marinegpt} for generating descriptions. 
The detected instances (highlighted in colored boxes) are individually processed to produce detailed textual descriptions, with separate captions generated for the entire image (image-level) and each detected object (instance-level).}
\label{fig:sample1}
\end{figure*}

Fig. \ref{fig:sample1} illustrate the processes of image-level and instance-level caption generation, utilizing MRegionCLIP for object instance detection and MarineGPT for caption generation \cite{zheng2023marinegpt}.

\begin{comment}
\begin{figure*}[t!]
\centering
\includegraphics[width=0.9\linewidth]{image_text_gen2.pdf}
\caption{Unsupervised image and instance levels captions generation processes on a sample image using object detector MRegionCLIP and MarineGPT \cite{zheng2023marinegpt}. }
\label{fig:sample2}
\end{figure*}
\end{comment}

\begin{comment}
\begin{figure*}[t!]
\centering
\includegraphics[width=0.9\linewidth]{image_text_gen3.pdf}
\caption{Unsupervised image and instance levels captions generation processes on a sample image using object detector MRegionCLIP and MarineGPT \cite{zheng2023marinegpt}. }
\label{fig:sample3}
\end{figure*}
\end{comment}

\section{Underwater Datasets and Tasks}
\label{datasets}
\subsection{Zero-shot Marine Species Classification Datasets}
For zero-shot classification tasks, we used the following datasets:

\begin{enumerate}
    \item \textbf{Marine Animal Images Dataset \cite{marine_animal}}: This dataset contains 806 images across nine different categories of sea animals, including Fish, Goldfish, Harbor Seal, Jellyfish, Lobster, Oyster, Sea Turtle, Squid, and Starfish. 
The images vary in size, lighting conditions, backgrounds, and camera angles, with an average resolution of $1024 \times 768$.
The dataset is split into 621 training images and 185 test images.

\item \textbf{Sea Animals Images Dataset \cite{sea_animal}}: This dataset includes 13,711 images covering 23 different sea animals, such as Clams, Corals, Crabs, Dolphins, Eels, Fish, Jellyfish, Lobsters, Nudibranchs, Octopuses, Otters, Penguins, Puffers, Sea Rays, Sea Urchins, Seahorses, Seals, Sharks, Shrimps, Squids, Starfish, Turtles, and Whales, with an average resolution of $300 \times 225$.
\end{enumerate}

\subsection{Zero-shot Fine-Grained Classification Datasets}
For these experiments, we used the following datasets:

\begin{enumerate}
    \item \textbf{FishNet \cite{khan2023fishnet}:} This large-scale, diverse dataset contains 94,532 images of fishes from 17,357 aquatic species, organized according to biological taxonomy (order, family, genus, and species). 
The dataset covers 83 orders and 463 families, with an average resolution of $542 \times 372$. 
We report classification performance at both the family and order levels.

\item \textbf{FishNet Open Images dataset \cite{kay2021fishnet}:} This dataset consists of 86,029 images across 34 object classes, representing a large and diverse public dataset of fisheries with an average resolution of $1261 \times 722$.
It poses several challenges, including visual similarity between species, imbalanced class distributions, harsh weather conditions, and chaotic crew activities.

\item \textbf{Large Scale Fish Dataset \cite{ulucan2020large}:} This dataset includes images of 9 different types of seafood collected from a supermarket: gilt-head bream, red sea bream, sea bass, red mullet, horse mackerel, black sea sprat, striped red mullet, trout, and shrimp. 
Images have an average resolution of $1688 \times 1267$. 
The dataset contains a total of 18,000 images, with 2,000 samples per class after data augmentation

\end{enumerate}

\subsection{Zero-shot Coral Species Classification Datasets}
This experiment was conducted using the following two datasets:

\begin{enumerate}
    \item \textbf{Coral Species Classification (CSC) Dataset \cite{coral-species-classification_dataset}:} This dataset contains 16 coral species classes with a total of 202 images, averaging a resolution of $257 \times 192$. 
The classes are Acanthastrea-echinata, Acropora-millepora, Astreopora-myriophthalma, Coscinaraea-columna, Cyphastrea-microphthalma, Diploastrea-heliopora, Favites-pentagona, Goniopora-lobata, Montipora-stellata, Pavona-cactus, Plesiastrea-versipora, Pocillopora-damicornis, Porites-lobata, Psammocora-contigua, Stylophora-pistillata, and Turbinaria-peltata.

\item \textbf{Corals Classification (CC) Dataset \cite{coral_classification}:} This dataset consists of 9,292 images categorized into two classes: Healthy and Bleached Corals. 
The images have an average resolution of $294 \times 231$ and are divided into training, testing, and validation splits in an 80-10-10 ratio, resulting in 7,384 training images, 923 testing images, and 985 validation images.

\end{enumerate}

\subsection{Downstream Tasks and Datasets}
We evaluated our AquaticCLIP model on several downstream analysis tasks, including salient object segmentation, instance segmentation, semantic segmentation, object detection and classification, and object counting.

For supervised salient object segmentation on underwater images, we used the USOD10K dataset \cite{usod10k}.
For instance segmentation of marine images, we employed the Underwater Image Instance Segmentation (UIIS) dataset \cite{Lian_2023_ICCV}, and for semantic segmentation, we used the SUIM dataset \cite{islam2020suim}.
Fine-tuned underwater object detection and classification tasks were conducted with four datasets: FishNet \cite{khan2023fishnet}, DeepFish \cite{saleh2020realistic}, URPC \cite{urpc}, and Brackish \cite{Pedersen_2019_CVPR_Workshops}.
For marine object counting in biodiversity studies, we used the IOCFish5K dataset \cite{sun2023indiscernible}.

\begin{enumerate}
    \item \textbf{Instance, Salient Object, and Semantic Segmentation Datasets:} The fine-tuned segmentation experiments were performed on the following datasets:

\begin{enumerate}
    \item \textbf{Underwater Image Instance Segmentation (UIIS) \cite{Lian_2023_ICCV}:} This dataset contains 4,628 images across 7 categories with pixel-level annotations for underwater instance segmentation. 
Categories include fish, coral reefs, aquatic plants, and wrecks/ruins, which are crucial for marine exploration and ecological studies. 
The images vary in resolution from $240 \times 320$ to $720 \times 1280$.

\item \textbf{USOD10K \cite{usod10k}:} This dataset is used for underwater salient object segmentation and contains 10,255 underwater images with a resolution of $640 \times 480$ pixels, covering 70 categories of salient objects across 12 underwater scenes. 
The dataset is divided into 7,178 training images, 2,051 validation images, and 1,026 testing images.

\item \textbf{Semantic Segmentation of Underwater IMagery (SUIM) \cite{islam2020suim}:} 
This dataset includes 1,635 image-mask pairs across 8 object categories for segmentation, including human divers, aquatic plants and seagrass, wrecks and ruins, robots, reefs and invertebrates, fish and vertebrates, sea-floor and rocks, and background. 
The masks are available in binary and combined RGB formats. 
The dataset is split into 1,525 training and 110 validation samples, with an average image resolution of $294 \times 231$ pixels.
\end{enumerate}

 \begin{figure*}[t!]
\centering
\includegraphics[width=\linewidth]{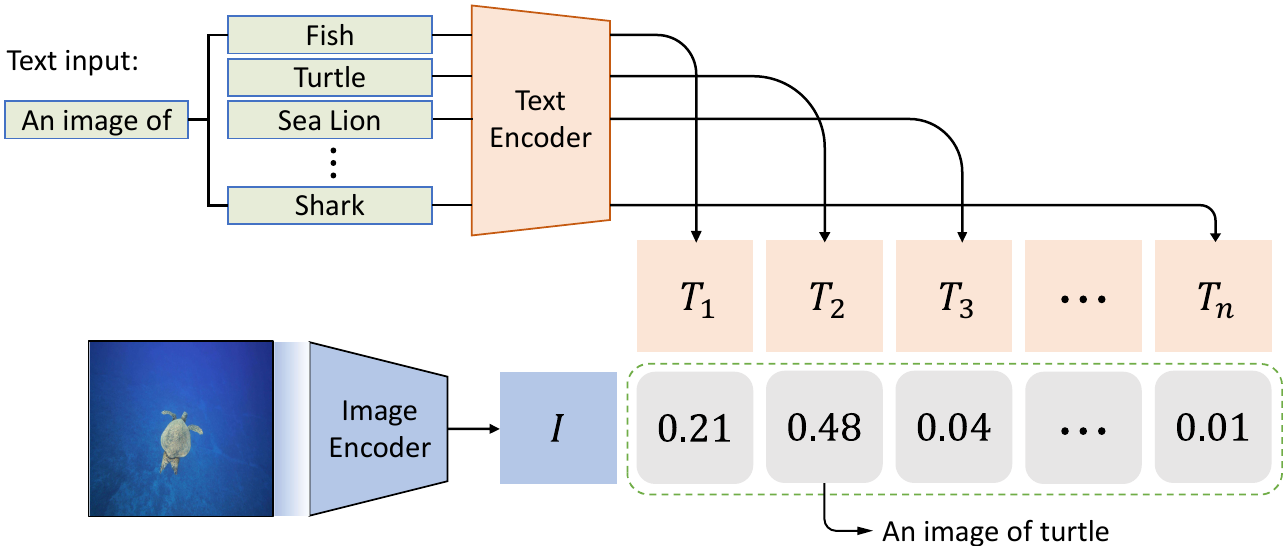}
\caption{Zero-shot inference using the proposed AquaticCLIP model. The process involves feeding a text input (e.g., ``An image of'') with various marine categories (e.g., Fish, Turtle, Sea Lion, Shark) into a text encoder to generate textual embeddings $T_{1}, T_{2}, T_{3},...,T_{n}$.
An input image is processed by the image encoder to produce an image embedding I.
Similarity scores between I and each $T_{1}, T_{2}, T_{3},...,T_{n}$ are calculated, with the highest score (e.g., 0.48 for Turtle) indicating the model’s prediction for the image's content.}
\label{fig:inference}
\end{figure*}

\item \textbf{Object Detection and Classification Datasets:} The object detection and classification tasks were performed on the following datasets:
\begin{enumerate}
    \item  \textbf{FishNet \cite{khan2023fishnet}:} Previously discussed in the zero-shot classification task section.
    \item \textbf{DeepFish Dataset \cite{saleh2020realistic}:} Contains 39,766 images for tasks such as localization, counting, segmentation, and classification, collected from 20 habitats in tropical Australian marine environments. 
The images are captured in full HD resolution ($1920 \times 1080$) from digital cameras. 
The dataset is split as follows: 19,883 images for training, 7,953 for validation, and 11,930 for testing in fish classification; 1,600, 640, and 960 for fish detection; and 310, 124, and 186 for fish segmentation.
\item \textbf{Brackish Dataset \cite{Pedersen_2019_CVPR_Workshops}:} Contains high-resolution ($1920 \times 1080$) images with bounding box annotations for 6 categories (big fish, small fish, starfish, shrimps, jellyfish, and crabs) captured in a brackish strait with variable visibility.
It consists of 14,518 images, divided into training, validation, and test sets with an 80$\%$:10$\%$:10$\%$ split.
\item \textbf{URPC Dataset \cite{urpc}:} This dataset contains 6,626 high-resolution real underwater images with an average resolution of $2,388 \times 1,345$ pixels. 
The images feature four types of underwater objects: holothurian, echinus, scallop, and starfish. 
Identifying and distinguishing these objects is challenging due to their varying sizes and their tendency to blend into the marine background, making this dataset particularly difficult for underwater object detection tasks.
    
\end{enumerate}

\item \textbf{Underwater Object Counting Dataset:}  This experiment uses the following dataset:

\begin{enumerate}
    \item \textbf{IOCfish5k \cite{sun2023indiscernible}:} This dataset includes 5,637 annotated images with an average resolution of 1080 × 1920 pixels. It contains a total of 659,024 object counts, with minimum, average, and maximum object counts per image being 0, 117, and 2,371, respectively.
\end{enumerate}
    
\end{enumerate}

\section{SOTA Methods for Comparison}
\label{sota}
We compared our AquaticCLIP model with a range of state-Of-The-Art (SOTA) methods across various underwater image analysis tasks.

\begin{enumerate}
    \item \textbf{For the zero-shot classification task}, we evaluated the performance of AquaticCLIP against existing vision-language models (VLMs), including Frozen CLIP \cite{radford2021learning}, Finetune CLIP \cite{radford2021learning}, and prompt-based VLMs such as CoOp \cite{zhou2022learning} and MAPLE \cite{khattak2023maple}, as well as GPT4V \cite{yang2023dawn}, BLIP2 \cite{li2023blip}, and MarineGPT \cite{zheng2023marinegpt}.
    CoOp and MAPLE were fine-tuned using the original source codes provided by their authors. 
    \item   \textbf{For the supervised classification task}, we compared AquaticCLIP with models like ResNet-34/50/101 \cite{he2016deep}, ViT-S/B/L \cite{alexey2020image}, BeiT \cite{bao2021beit}, ConvNeXt \cite{liu2022convnet}, ConvNeXt \cite{liu2022convnet} + Focal Loss (FL) \cite{lin2017focal}, and ConvNeXt \cite{liu2022convnet} + Class-Balanced (CB) \cite{cui2019class}.
  To ensure a fair comparison, we followed the settings used in FishNet \cite{khan2023fishnet}.
\item In the \textbf{underwater salient object segmentation task}, AquaticCLIP was evaluated against SOTA methods such as DUAL-SAM \cite{zhang2024fantastic}, SVAM-Net \cite{jahidul2020svam}, CTDNet \cite{zhao2021complementary}, CDINet \cite{zhang2021cross}, SGL-KRN \cite{xu2021locate}, TC-USOD \cite{usod10k}, and MarineInst \cite{ziqiang2024marineinst}.
\item \textbf{For underwater instance segmentation}, we compared it with methods like Point Rend \cite{pointrend_Kirillov_2020_CVPR}, SOLOv2 \cite{solov2_Wang2020}, QueryInst \cite{query_inst_Fang_2021_ICCV}, Mask Transformer \cite{mask_transfiner_Ke_2022_CVPR}, Mask2Former \cite{cheng2021mask2former}, WaterMask R-CNN \cite{sun2023indiscernible}, and Cascade WaterMask R-CNN \cite{sun2023indiscernible}. 
\item In the \textbf{underwater semantic segmentation task}, AquaticCLIP was compared to models including PSPNet \cite{p2pnet_Song_2021_ICCV}, DeepLabv3 \cite{deeplab}, SUIM-Net \cite{islam2020suim}, and  Mask2Former \cite{cheng2021mask2former}.
\item \textbf{For object detection and classification tasks}, we benchmarked AquaticCLIP against FasterRCNN \cite{ren2016faster}, YOLOF \cite{chen2021you}, TOOD \cite{feng2021tood}, MarineInst \cite{ziqiang2024marineinst}, MarineDet \cite{haixin2023marinedet}, and MRegionCLIP.
\item \textbf{For object counting tasks}, we compared it with existing SOTA methods, including KDMG \cite{kdmg}, MPS \cite{mps}, CLTR \cite{cltr_liang2022end}, and IOCFormer \cite{sun2023indiscernible}. 
All methods were implemented using the official source codes provided by the original authors and evaluated on the same datasets for consistency.
\end{enumerate}

\section{SOTA Methods Training Details}
\label{sotadetails}
\begin{itemize}
    \item \textbf{Supervised Classification Methods:} For a fair comparison, we used both ResNet-based and ViT-based architectures, including ResNets (ResNet-34, ResNet-50, ResNet-101) \cite{he2016deep}, ViT (ViT-S, ViT-B, ViT-L) \cite{alexey2020image}, BeiT \cite{bao2021beit}, and ConvNeXt \cite{liu2022convnet}.
The classification head was replaced with two fully connected (FC) layers, with a dropout rate of 0.5 applied to the first FC layer.
All backbone networks were pre-trained on ImageNet and fine-tuned using the Adam optimizer over 100 epochs on each classification dataset's training splits. 
The initial learning rates were set to $3 \times 10^{-5}$ for the backbone networks and $3 \times 10^{-4}$ for the classification heads, with the learning rate halved every 20 epochs. 
To address the long-tail classification issue, we also evaluated focal loss \cite{lin2017focal} and class-balanced training \cite{cui2019class} techniques and compared classification performance. 
For FishNet, all results are reported for fish family classification.
\item \textbf{Supervised Object Detection and Classification Methods:} We evaluated five SOTA methods for object detection and classification: FastRCNN \cite{ren2016faster}, YOLOF \cite{chen2021you}, and TOOD \cite{feng2021tood}. The backbone networks for these methods were pre-trained on the MS-COCO dataset \cite{lin2014microsoft}, and we fine-tuned the complete models using the stochastic gradient descent (SGD) optimizer with a learning rate of $2.5 \times 10^{-4}$,, momentum of 0.9, and weight decay of $1 \times 10^{-4}$ for 80 epochs. 
Performance was reported on the official training and testing splits of the evaluated datasets to ensure fair comparison.

\end{itemize}

\section{Evaluation Metrics}
\label{metrics}
We used various evaluation metrics suited to the specific aquatic imagery tasks. 

\begin{itemize}
    \item For \textbf{zero-shot classification tasks}, we reported accuracy and $F_{1}$ scores.
    \item For \textbf{zero-shot cross-modal retrieval tasks}, we evaluated performance using R$@$1, R$@$50, and R$@$200.
    \item In the \textbf{underwater salient object segmentation task}, metrics included S-measure ($S_{m}$), max E-measure ($E_{\epsilon}^{max}$), max F-measure (max), and Mean Absolute Error (MAE) \cite{usod10k}.
    \item For \textbf{instance segmentation} on the UIIS dataset, we used mAP, $\textrm{AP}_{50}$, $\textrm{AP}_{75}$, $\textrm{AP}_{S}$, $\textrm{AP}_{M}$,  $\textrm{AP}_{L}$, $\textrm{AP}_{f}$, $\textrm{AP}_{h}$, and $\textrm{AP}_{r}$ metrics \cite{Lian_2023_ICCV}.
    \item \textbf{Semantic segmentation} on the SUIM dataset was evaluated with IoU and $F$-measure scores.
    \item \textbf{Underwater object detection and classification tasks} are evaluated using mAP$_{50}$ metric.
    \item For \textbf{underwater object counting tasks}, we assessed performance using MAE and MSE metrics.
\end{itemize}

\section{Zero-shot Inference}
\label{inference}
Fig. \ref{fig:inference} illustrates the zero-shot inference process, where normalized image and text embeddings are estimated and used for cosine similarity to assign class labels.

\section{Zero-shot Cross-Modal Retrieval Results}
\label{cross}
We evaluated zero-shot text-to-image and image-to-text retrieval tasks by finding the closest matches for each modality. 
Metrics such as R$@$1, R$@$50, and R$@$200 were used to check if the correct ground-truth pair was among the top matches. 
Table \ref{table4} presents zero-shot cross-modal retrieval results on a dataset of 2K image-text pairs (excluded from the main 2M dataset), along with comparisons to four SOTA VLMs. 
AquaticCLIP outperformed all other methods by a significant margin, showcasing a strong alignment of cross-modal features across diverse visual and textual domains. 
MarineGPT ranked as the second-best performer.

\section{Underwater Object Detection and Classification Results}
\label{objdetect}
For object detection and classification, we replaced ResNet-50 in MRegionCLIP with our pre-trained image encoder $\Psi_{v}$, naming the model AquaticDet. 
We applied the same fine-tuning settings as previously described. Table \ref{table_ood} shows object detection and classification results across four datasets, with AquaticDet achieving the highest mAP$_{50}$ scores across all datasets. 
This superior performance is attributed to pre-training on the 2M image-text paired dataset, which enabled AquaticDet to extract highly efficient visual features. 
The addition of the prompt-guided vision encoder and vision-guided text encoder, along with comprehensive captions at both image and instance levels, contributed to significant performance improvements even in challenging conditions.

\begin{table}[t!]
\caption{Comparison of supervised object detection and classification results (mAP$_{50}$) across four datasets: FishNet, DeepFish, Brackish, and URPC. 
AquaticDet achieves the highest mAP$_{50}$ scores across all datasets, outperforming other SOTA methods. 
Specifically, AquaticDet scores 0.903 on FishNet, 0.891 on DeepFish, 0.877 on Brackish, and 0.837 on URPC. 
For FishNet, the performance is reported for common family classes.}
\begin{center}
\makebox[\linewidth]{
\scalebox{0.90}{
\begin{tabu}{|c|c|c|c|c|}
\tabucline[1.5pt]{-}
Methods&FishNet&DeepFish&Brackish&URPC\\\tabucline[0.5pt]{-}
FasterRCNN \cite{ren2016faster}&0.284&0.814&0.788&0.475\\
YOLOF \cite{chen2021you}&0.672&0.806&0.813&0.511\\
TOOD \cite{feng2021tood}&0.811&0.766&0.805&0.507\\
MRegionCLIP&\underline{0.867}&0.855&\underline{0.842}&0.758\\
MarineInst \cite{ziqiang2024marineinst}&\underline{0.868}&0.854&0.841&\underline{0.779}\\
MarineDet \cite{haixin2023marinedet}&-&-&-&0.706\\
AquaticDet &\textbf{0.903}&\textbf{0.891}&\textbf{0.877}&\textbf{0.837}\\\tabucline[0.5pt]{-}
%DeepLabv3 \cite{deeplab}&0.791&0.812\\
%SUIM-Net \cite{islam2020suim}&0.841&0.869\\
%Mask2Former \cite{cheng2021mask2former}&\underline{0.855}&\underline{0.896}\\
%AquaticSeg&\textbf{0.881}&\textbf{0.921}\\\tabucline[0.5pt]{-}
\end{tabu}
}}
\end{center}
\label{table_ood}
\end{table}

\begin{table}[t!]
\caption{Supervised salient object segmentation results on the USOD10K dataset \cite{usod10k}.
AquaticSAM outperforms other methods, including SVAM-Net \cite{jahidul2020svam}, CTDNet \cite{zhao2021complementary}, CDINet \cite{zhang2021cross}, SGL-KRN \cite{xu2021locate}, TC-USOD \cite{usod10k}, and MarineInst \cite{ziqiang2024marineinst}, achieving the best scores for $S_{m}$ (0.943), $E^{max}_{\epsilon}$ (0.978), $\max F$ (0.942), and the lowest MAE (0.012), despite not using additional cues such as depth or edge information.}
\begin{center}
\makebox[\linewidth]{
\scalebox{0.90}{
\begin{tabu}{|c|c|c|c|c|}
\tabucline[1.5pt]{-}
Methods&$S_{m} \uparrow$&$E^{max}_{\epsilon} \uparrow$&$\max F \uparrow$&MAE $\downarrow$\\\tabucline[0.5pt]{-}
SVAM-Net \cite{jahidul2020svam}&0.746&0.764&0.645&0.091 \\
CTDNet \cite{zhao2021complementary}&0.908&0.953&0.907&0.028\\
CDINet \cite{zhang2021cross}&0.704&0.864&0.736&0.090\\
SGL-KRN \cite{xu2021locate}&0.921&0.963&0.924&0.023\\
TC-USOD \cite{usod10k}&0.921&\underline{0.968}&0.923&0.020\\
DUAL-SAM \cite{zhang2024fantastic}&\underline{0.923}&\underline{0.968}&\underline{0.931}&\underline{0.018}\\
MarineInst \cite{ziqiang2024marineinst}&0.910&0.941&0.887&0.025\\
AquaticSAM&\textbf{0.943}&\textbf{0.978}&\textbf{0.942}&\textbf{0.012}\\\tabucline[0.5pt]{-}
\end{tabu}
}}
\end{center}
\label{table_sos}
\end{table}

\begin{table*}[t!]
    \centering
   \caption{Comparison of instance segmentation results on the UIIS dataset \cite{Lian_2023_ICCV} between AquaticSAM and other SOTA methods, all using ResNet101 as the backbone. 
AquaticSAM achieves the highest mAP (0.293), AP\textsubscript{50} (0.451), and AP\textsubscript{\textit{h}} (0.576), demonstrating superior performance in small object segmentation and overall accuracy.}
   \adjustbox{width=0.95\textwidth}{
    \begin{tabular}{ l c| c c | c c c | c c c |}
    \hline
    \multicolumn{1}{c}{Method} & mAP $\uparrow$ & AP\textsubscript{50} $\uparrow$ & AP\textsubscript{75} $\uparrow$ & AP\textsubscript{\textit{S}} $\uparrow$ & AP\textsubscript{\textit{M}} $\uparrow$ & AP\textsubscript{\textit{L}} $\uparrow$ & AP\textsubscript{\textit{f}} $\uparrow$ & AP\textsubscript{\textit{h}} $\uparrow$ & AP\textsubscript{\textit{r}} $\uparrow$  \\
    \hline
   Point Rend \cite{pointrend_Kirillov_2020_CVPR} & 0.259 & 0.434 & 0.276 & 0.820 & 0.202 & 0.386 & 0.433 & 0.541 & 0.206 \\
   SOLOv2 \cite{solov2_Wang2020} & 0.245 & 0.409 & 0.251 & 0.560 & 0.194 & 0.376 & 0.364 & 0.483 & 0.206  \\
   QueryInst \cite{query_inst_Fang_2021_ICCV}  & 0.260 & 0.428 & 0.273 & 0.820 & 0.217 & 0.351 & 0.433 & 0.541 & 0.206 \\
   Mask Transfiner \cite{mask_transfiner_Ke_2022_CVPR}  & 0.246 & 0.421 & 0.260 & 0.720 & 0.194 & 0.361 & 0.438 & 0.263 & 0.198  \\
   Mask2Former \cite{cheng2021mask2former}  & 0.257 & 0.380 & 0.277 & 0.630 & 0.189 & 0.381 & 0.411 & 0.519 & 0.231  \\
   WaterMask R-CNN \cite{sun2023indiscernible}  & \underline{0.272} & \underline{0.437} & 0.293 & \textbf{0.900} & \underline{0.218} & \underline{0.389} & 0.463 & 0.548 & 0.209 \\
   Cascade WaterMask R-CNN \cite{sun2023indiscernible}  & 0.271 & 0.429 & \underline{0.304} & 0.830 & 0.210 & \underline{0.389} & \underline{0.470} & \underline{0.558} & \underline{0.225} \\
   MarineInst \cite{ziqiang2024marineinst} & 0.266& 0.434& 0.298 & 0.832& 0.183& 0.367& 0.450 & 0.521 & 0.198\\
   AquaticSAM &\textbf{0.293}& \textbf{0.451} & \textbf{0.330} & \underline{0.870} & \textbf{0.236} & \textbf{0.403} & \textbf{0.513} & \textbf{0.576} & \textbf{0.252} \\
       \hline
    \end{tabular}}
    \label{table_is}
\end{table*}

\section{Underwater Scene Segmentation Results}
\label{segmentation}
For the supervised segmentation task, we used the SAM foundational model \cite{Kirillov_2023_ICCV_SAM}, which consists of a prompt encoder, an image encoder, and a lightweight mask decoder. 
In our fine-tuning experiments, we replaced SAM’s original image encoder with the AquaticCLIP vision encoder $\Psi_{v}$, tuned explicitly for aquatic scenes. 
Other settings and implementation details followed SAM’s original setup. 
We named our instance segmentation model AquaticSAM.

\begin{enumerate}
    \item \textbf{Salient Object Segmentation Results:} Table \ref{table_sos} shows the results of salient object segmentation on the USOD10K \cite{usod10k} dataset, where AquaticSAM consistently outperformed six existing SOTA methods across all metrics without relying on additional information such as edge or depth maps. 
While MarineInst also performed well, it lagged behind AquaticSAM. 

\item \textbf{Underwater Instance Segmentation Results:} In Table \ref{table_is}, results for underwater instance segmentation on the UIIS \cite{Lian_2023_ICCV} dataset indicate that AquaticSAM achieved the highest mAP, AP$_{50}$, and AP$_{75}$ scores, though it ranked second-best for small object segmentation (APS), with WaterMask R-CNN, its cascaded variant, and MarineInst also performing strongly.

\item \textbf{Semantic Segmentation Results (Table \ref{table_ss}):} For semantic segmentation, we introduced AquaticSeg, which incorporates a classification head fine-tuned for pixel-wise classification. 
Following the experimental protocols of MarineInst \cite{ziqiang2024marineinst}, these segmentation experiments used the training splits for each dataset. 
Table \ref{table_ss} presents the semantic segmentation results on the SUIM \cite{islam2020suim} dataset, where AquaticSeg achieved the highest mIOU and F-score, with Mask2Former and MarineInst as the second- and third-best performers.
\end{enumerate}

\noindent Overall, both AquaticSAM and AquaticSeg benefited from pre-training on the 2M aquatic images dataset. 
Additionally, the inclusion of both instance-level and image-level captions for language supervision significantly improved performance compared to using only image-level captions.

\begin{table}[t!]
\caption{Comparison of semantic segmentation results on the SUIM \cite{islam2020suim} dataset between AquaticSeg and other SOTA methods. 
AquaticSeg achieves the highest mIoU (0.881) and F-score (0.921), demonstrating superior performance in underwater semantic segmentation tasks.}
\begin{center}
\makebox[\linewidth]{
\scalebox{0.90}{
\begin{tabu}{|c|c|c|}
\tabucline[1.5pt]{-}
Methods&mIoU $\uparrow$&F-score $\uparrow$\\\tabucline[0.5pt]{-}
PSPNet \cite{p2pnet_Song_2021_ICCV}& 0.774 & 0.760\\
DeepLabv3 \cite{deeplab}&0.791&0.812\\
SUIM-Net \cite{islam2020suim}&0.841&0.869\\
Mask2Former \cite{cheng2021mask2former}&\underline{0.855}&\underline{0.896}\\
MarinsInst \cite{ziqiang2024marineinst}&0.851&0.882\\
AquaticSeg&\textbf{0.881}&\textbf{0.921}\\\tabucline[0.5pt]{-}
\end{tabu}
}}
\end{center}
\label{table_ss}
\end{table}

\section{Results of Object Counting in Underwater Scenes (Table \ref{table5})}
\label{counting}
For object counting in underwater scenes, we employed a crowd localization transformer method \cite{cltr_liang2022end}, which includes a CNN-based backbone, a transformer encoder, a transformer decoder, and a nearest-neighbors matching component. 
In our experiments, we replaced the original backbone and encoder with the AquaticCLIP pre-trained vision encoder $\Psi_{v}$, specifically fine-tuned for aquatic environments.
The remaining settings followed the original implementation.
We refer to our object counting model as AquaticOC.

Table \ref{table5} presents object counting results on the IOCFish5K dataset \cite{sun2023indiscernible}, comparing AquaticOC with existing SOTA methods, including KDMG \cite{kdmg}, MPS \cite{mps}, CLTR \cite{cltr_liang2022end}, and IOCFormer \cite{sun2023indiscernible}.
All methods were implemented using the official source codes provided by the authors and evaluated on the same training and testing splits for fair comparison. 
AquaticOC achieved the best results in terms of MAE and MSE, showing a significant improvement over the baseline CLTR model, representing a strong advancement in efficient and accurate object counting for aquatic scenes.

\begin{table}[t!]
\centering
\caption{Zero-shot cross-modal retrieval results (text-to-image and image-to-text) on the 2K dataset, presented as (text-to-image $\%$ $|$ image-to-text $\%$).
AquaticCLIP outperforms other models, achieving the highest scores across R$@$1, R$@$50, and R$@$200, with top results of 22.29$\%$ $|$ 21.91$\%$ at R@1, 62.31$\%$ $|$ 63.20$\%$ at R@50, and 71.10$\%$ $|$ 72.34$\%$ at R@200.}
\begin{tabular}{|c | c| c| c| c| c| c|}
\hline
Methods & \multicolumn{3}{c|}{Our Dataset}  \\ \hline
& R@1 & R@50 &R@200 \\
Frozen CLIP&0.09$|$0.06&3.82$|$3.72&10.31$|$11.12\\
Finetune CLIP&\underline{13.98}$|$\underline{12.79}&43.42$|$42.31&61.34$|$60.74\\
GPT4V&8.19$|$8.17&35.10$|$\underline{34.13}&40.61$|$41.12\\
MarineGPT&\underline{20.12}$|$\underline{21.74}&\underline{60.78}$|$\underline{61.71}&\underline{69.58}$|$\underline{70.64}\\
AquaticCLIP&\textbf{22.29}$|$\textbf{21.91}&\textbf{62.31}$|$\textbf{63.20}&\textbf{71.10}$|$\textbf{72.34}\\
\hline
\end{tabular}
\label{table4}
%\vspace{-.7cm}
\end{table}

\section{Supervised AquaticCLIP Model: Linear Probe Evaluations}
\label{supervised}
We conducted linear probe experiments to evaluate the quality of visual features extracted by our AquaticCLIP model. 
Linear probing involves training a simple linear classifier, such as logistic regression, on top of features extracted from a pre-trained network, without fine-tuning the network’s original weights. 
In this experiment, we applied logistic regression on pre-extracted features from the AquaticCLIP vision encoder to assess the discriminative power and quality of the representations.

For downstream classification tasks across various marine vision datasets, we used logistic regression, following practices recommended by the self-supervised learning community \cite{geiping2023cookbook, caron2021emerging}.
We set the $\ell_{2}$ regularization coefficient $\lambda$ to $\frac{100}{MC}$, where $M$ is the embedding dimension and $C$ is the number of classes, and used the L-BFGS solver with a maximum of 1,000 iterations \cite{zhu1997algorithm}.
All downstream classification tasks were evaluated using stratified train-test splits or official splits when available.

\begin{table}[t!]
    \centering
   \caption{Object counting performance comparison of various SOTA methods on the IOCFish5k test set \cite{sun2023indiscernible}.
AquaticOC achieves the best results, with the lowest Mean Absolute Error (MAE: 13.50) and Mean Squared Error (MSE: 36.10), outperforming other methods.}
   \adjustbox{width=0.4\textwidth}{
    \begin{tabular}{ l| c c }
    \hline
    \multicolumn{1}{c}{Method} & MAE $\downarrow$ & MSE $\downarrow$  \\
    \hline
  %  MCNN \cite{mcnn_Zhang_2016_CVPR} & 72.93 & 129.43 & 4.90 \\
    %CSRNet \cite{csrnet_Li_2018_CVPR} & 38.12 & 69.75 & 2.48 \\
    %LCFCN \cite{lcfcn_Laradji_2018_ECCV} & 28.05 & 68.24 & 1.12 \\
    %CAN \cite{can_Liu_2019_CVPR} & 42.02 & 74.46 & 2.58 \\
    %DSSI-Net \cite{dssi_Liu_2019_ICCV} & 31.04 & 69.11 & 1.68 \\
    %BL \cite{bl_Ma_2019_ICCV} & 20.03 & 46.08 & 0.55 \\
    %NoisyCC \cite{noisycc} & 19.73 & 46.85 & 0.46 \\
    %DM-Count \cite{dmcount_Wang2020a} & 19.52 & 45.52 & 0.55 \\
    %GL \cite{gl_Wan_2021_CVPR} & 18.80 & 46.19 & 0.47 \\
    %P2PNet \cite{p2pnet_Song_2021_ICCV} & 20.74 & 47.90 & 0.48 \\
    KDMG \cite{kdmg} & 0.227 & 0.499  \\
    MPS \cite{mps} & 0.335 & 0.550  \\
    %MAN \cite{man_Liu2020} & 25.82 & 45.82 & 3.16 \\
    CLTR \cite{cltr_liang2022end} & 0.180 & 0.419  \\
    IOCFormer \cite{sun2023indiscernible} &\underline{ 0.171} & \underline{0.412}  \\
    AquaticOC&\textbf{0.135}&\textbf{0.361} \\
    \hline
        \end{tabular}}
    \label{table5}
\end{table}

\section{AquaticVision: Vision-Only Model Pre-training Details}
\label{vision}
For comparison, we also developed a vision-only model named AquaticVision, pre-trained using a self-supervised learning approach. 
For large-scale visual pre-training on our 2M aquatic imagery dataset, we employed DINOv2 \cite{oquab2023dinov2}, a SOTA self-supervised learning method based on student-teacher knowledge distillation for pre-training large ViT architectures.

We used a ViT-B architecture with 12 layers, 12 heads, an $8 \times 8$ patch size, a feed-forward network layer as MLP, GELU activation, an embedding dimension of 768, and a dropout rate of 0.1 \cite{dosovitskiy2020image}.
The model was trained with the AdamW optimizer with hyperparameter $\beta$ set to (0.9, 0.999), a batch size of 3072, and a maximum of 125,000 iterations.

%DINOv2 is an extension of two previous methods, DINO \cite{caron2021emerging} and iBOT \cite{zhou2021ibot}, and uses two main loss objectives including self-distillation loss and masked image modeling loss 
%to achieve SOTA results in linear probe accuracy.

%The teacher is updated as an exponential moving average of previous iterations of the student.
%Masked image modeling using an online tokenizer, introduced in iBOT, involves strategically masking specific regions in an input image and training the model to predict the masked regions based on
%the remaining contextual information. 
%This approach captures high-level visual features and context, inspired by masked language modeling in BERT \cite{kenton2019bert}.

\section{Visual Results}
\label{visual}
\begin{figure}[t!]
\centering
\includegraphics[width=\linewidth]{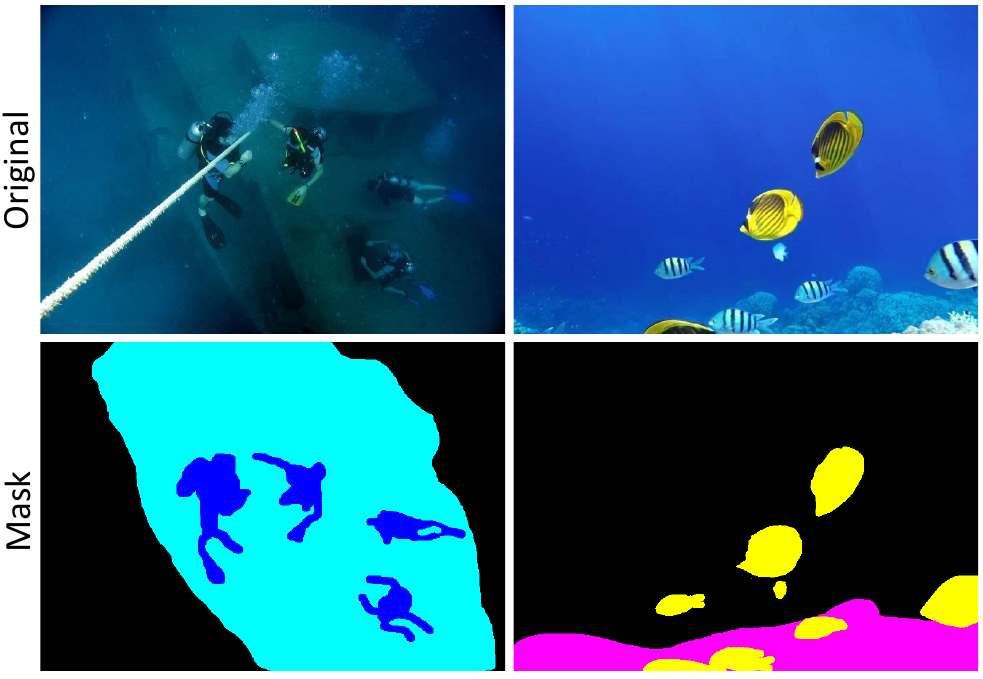}
\caption{ Semantic segmentation results on the SUIM dataset \cite{islam2020suim}.
 The top row shows original underwater images, while the bottom row displays corresponding segmentation masks generated by AquaticSeg, highlighting detected objects such as divers and fish with distinct color-coded regions.}
\label{fig:result_sem}
\end{figure}

\begin{figure}[t!]
\centering
\includegraphics[width=\linewidth]{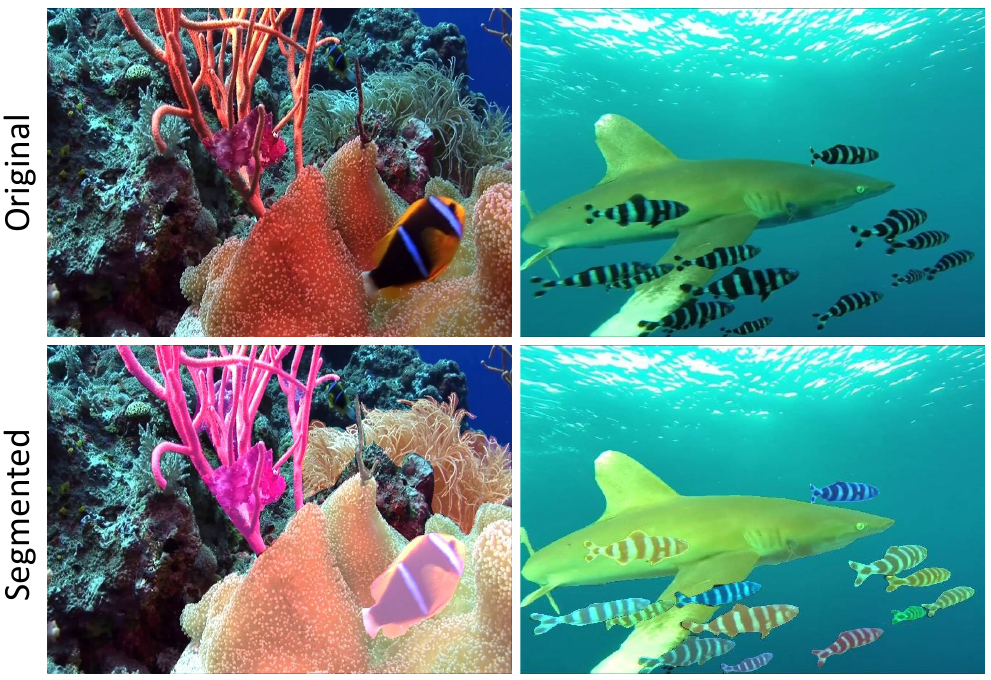}
\caption{Instance segmentation results on the UIIS dataset \cite{Lian_2023_ICCV}.
The top row shows original underwater scenes, while the bottom row presents segmented images generated by AquaticSAM, highlighting individual objects like fish, corals, and sharks with distinct boundaries for each detected instance.}
\label{fig:result_inst}
\end{figure}

\begin{figure}[t!]
\centering
\includegraphics[width=\linewidth]{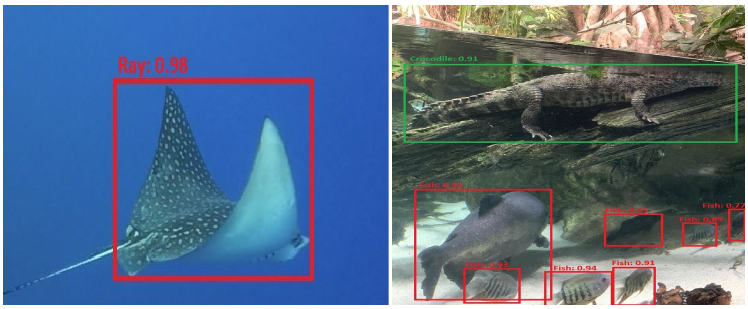}
\caption{Underwater object detection results on the FishNet dataset \cite{khan2023fishnet}.
The images show detected objects, including a ray, crocodile, and various fish, each marked with bounding boxes and confidence scores, demonstrating AquaticDet’s capability to accurately identify and label underwater species.}
\label{fig:result_det}
\end{figure}

\begin{figure}[t!]
\centering
\includegraphics[width=\linewidth]{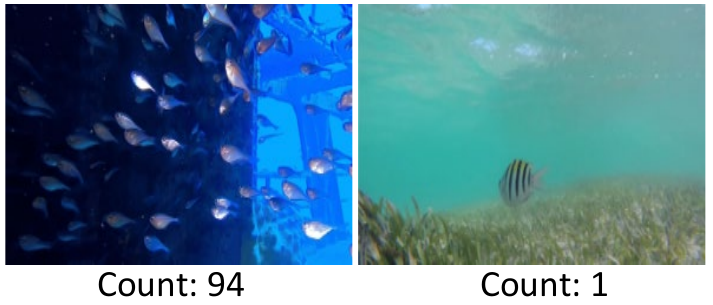}
\caption{Underwater object counting results on the IOCfish5k dataset \cite{sun2023indiscernible}.
The images show object counts generated by AquaticOC, with the left image counting 94 fish and the right image counting 1 fish, demonstrating the model’s ability to accurately detect and count objects in various underwater scenes.}
\label{fig:result_count}
\end{figure}

\begin{figure*}[t!]
\centering
\includegraphics[width=0.7\linewidth, height=0.6\linewidth]{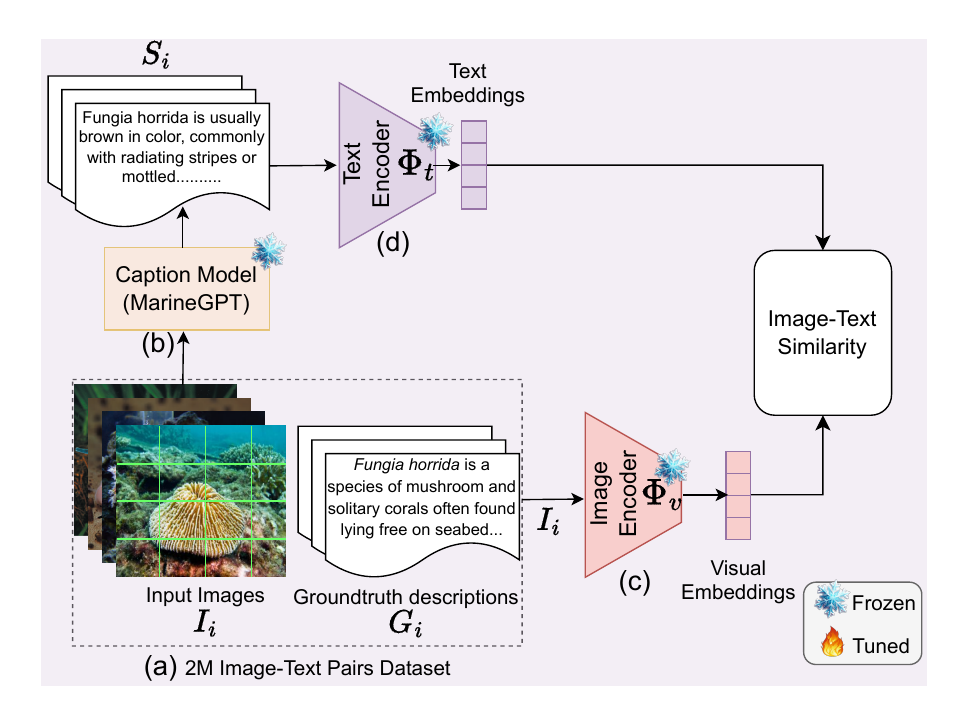}
\caption{Schematic illustration of the Frozen CLIP. Please refer to Table 1 in the main manuscript.}
\label{fig:frozen_clip}
\end{figure*}

\begin{figure*}[t!]
\centering
\includegraphics[width=\linewidth]{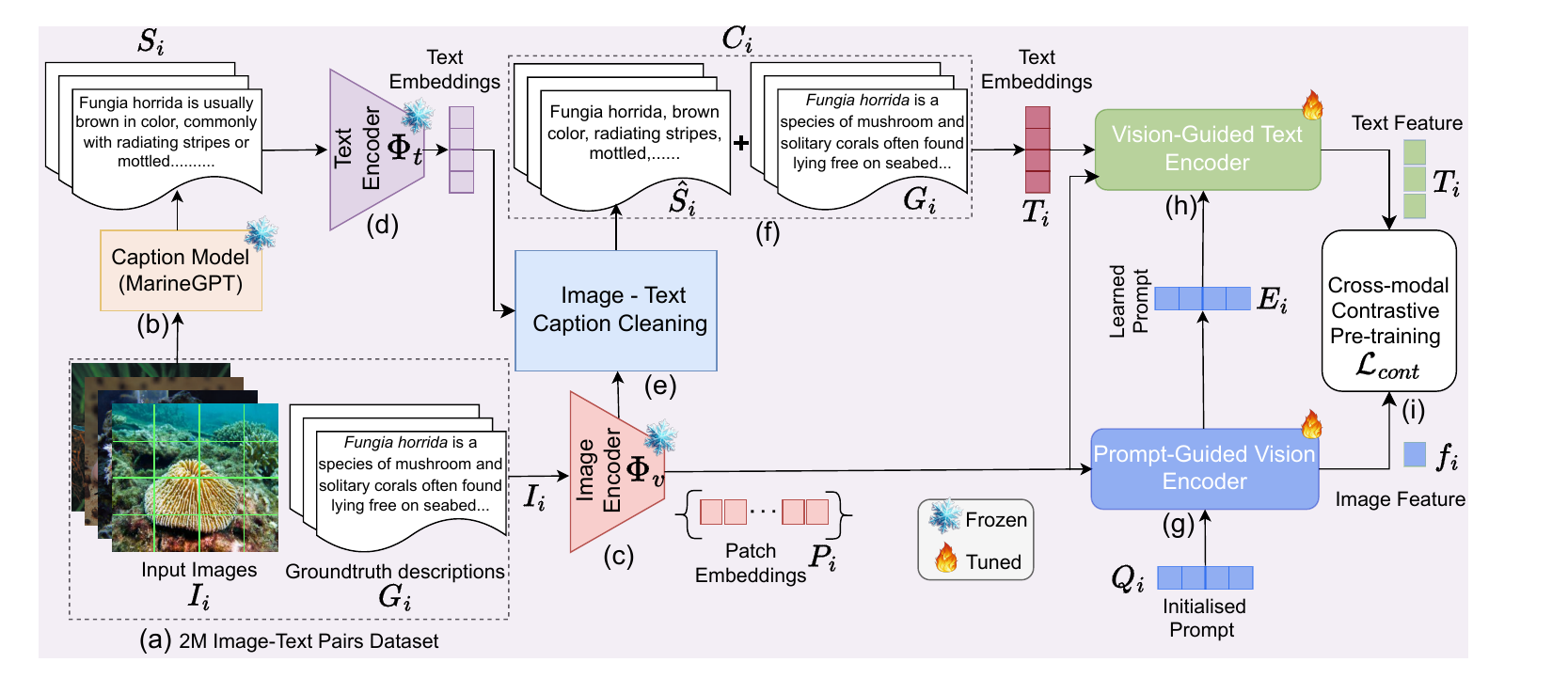}
\caption{Schematic illustration of the AquaticCLIP\textsubscript{1}. Please refer to Table 1 in the main manuscript.}
\label{fig:aquatic_clip1}
\end{figure*}

\begin{figure*}[t!]
\centering
\includegraphics[width=\linewidth, height=0.55\linewidth]{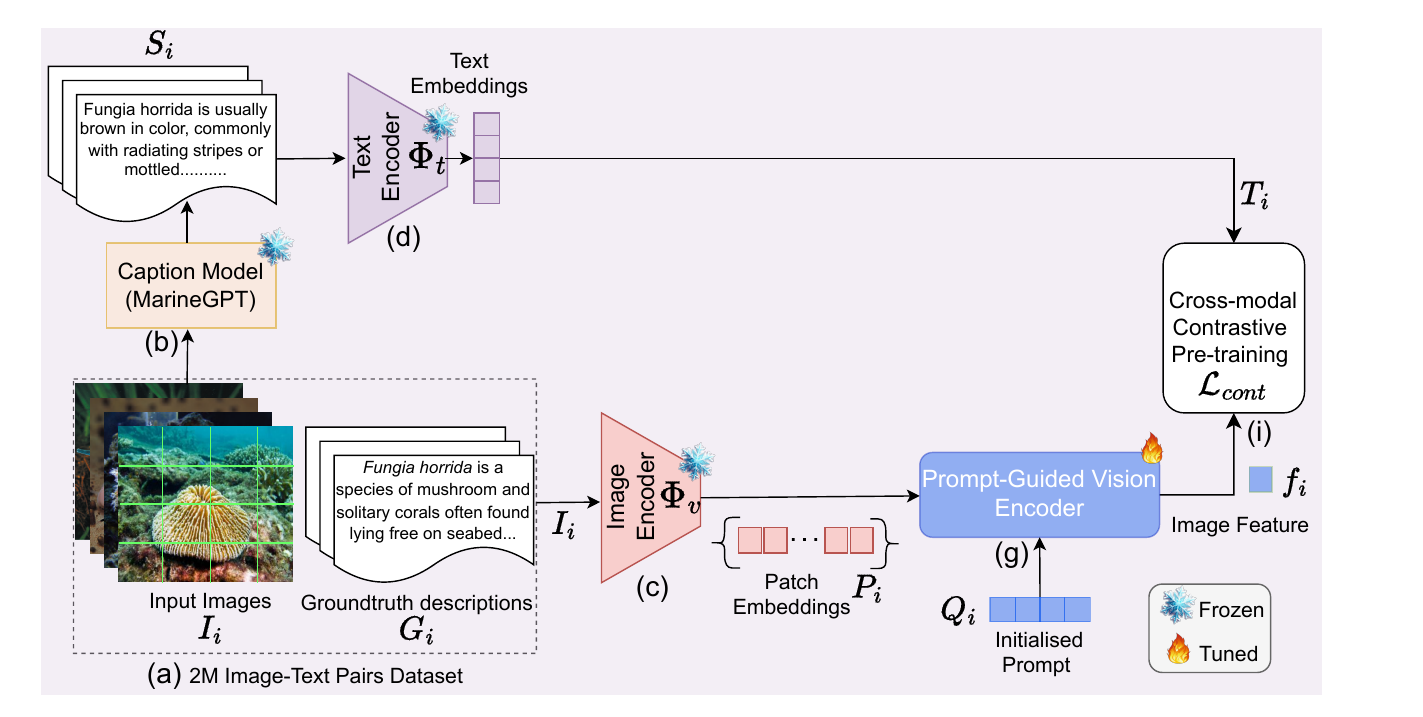}
\caption{Schematic illustration of the AquaticCLIP\textsubscript{2}. Please refer to Table 1 in the main manuscript.}
\label{fig:aquatic_clip2}
\end{figure*}

\begin{figure*}[t!]
\centering
\includegraphics[width=\linewidth]{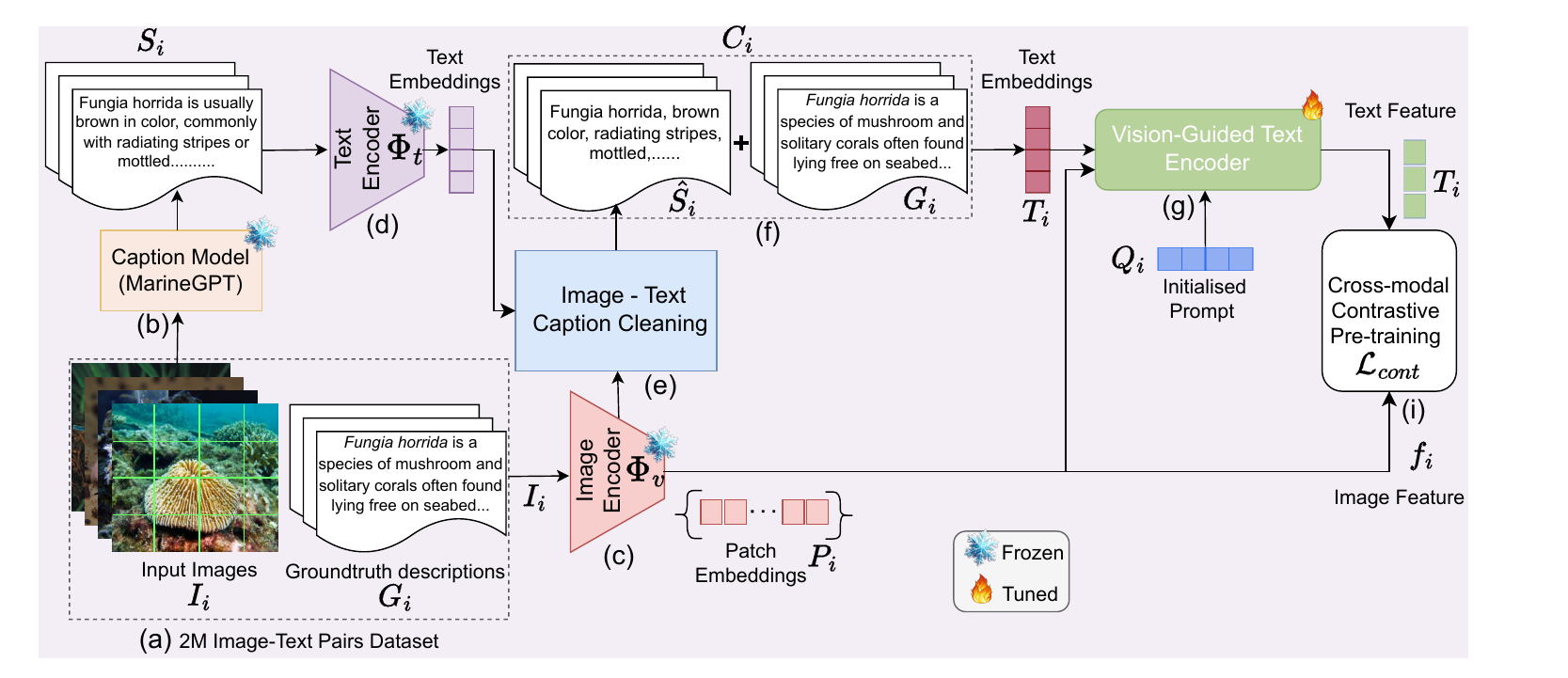}
\caption{Schematic illustration of the AquaticCLIP\textsubscript{3}. Please refer to Table 1 in the main manuscript.}
\label{fig:aquatic_clip3}
\end{figure*}

\begin{figure*}[t!]
\centering
\includegraphics[width=0.7\linewidth, height=0.6\linewidth]{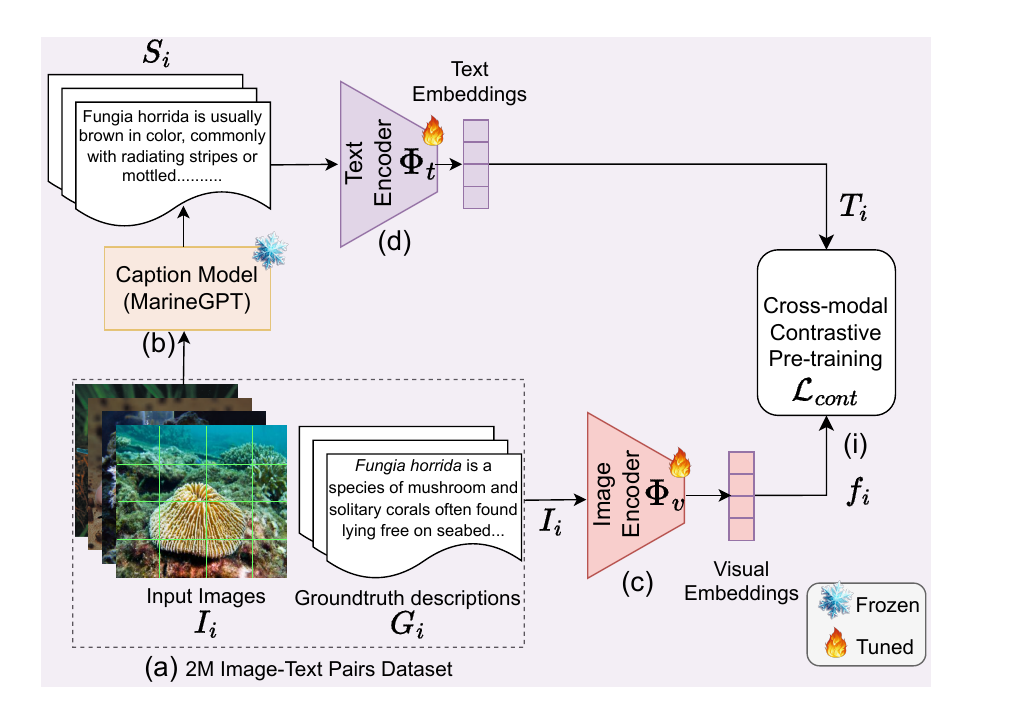}
\caption{Schematic illustration of the Finetuned CLIP. Please refer to Table 1 in the main manuscript.}
\label{fig:finetune_clip}
\end{figure*}

\begin{figure*}[t!]
\centering
\includegraphics[width=\linewidth]{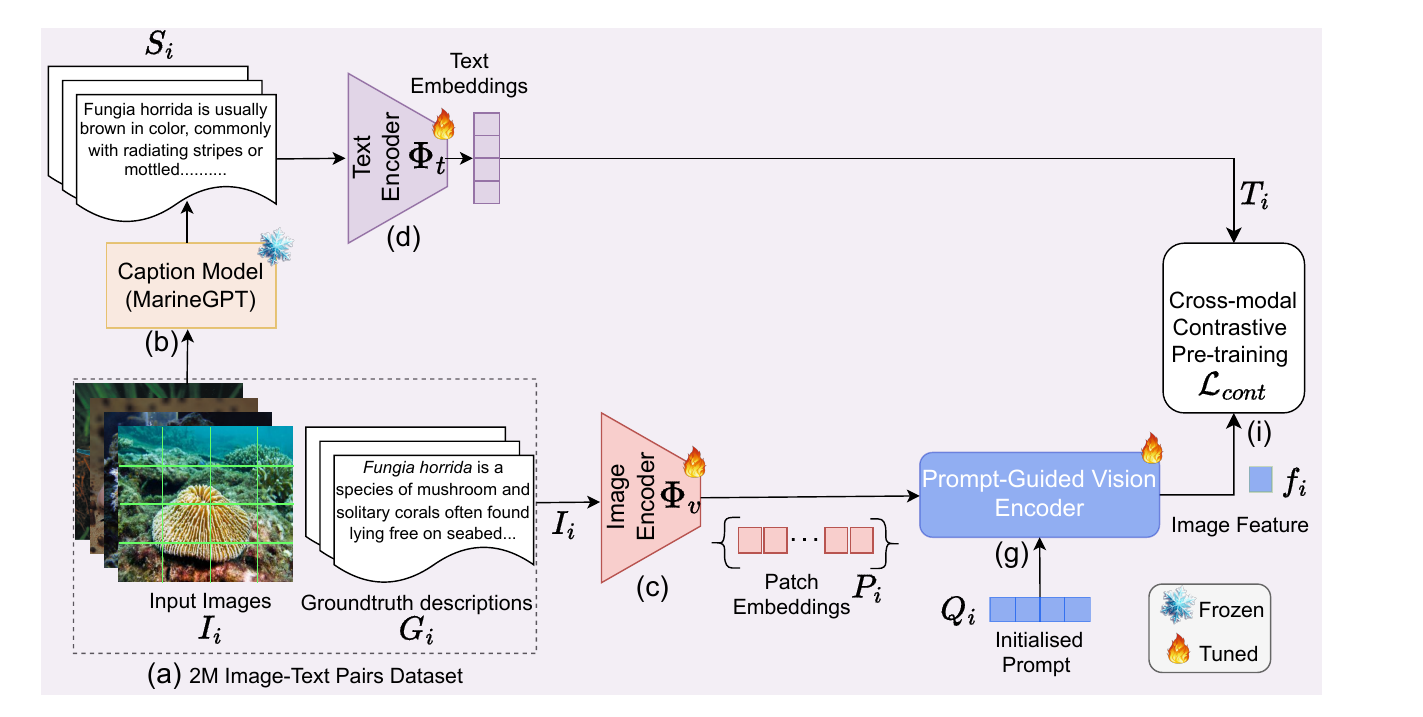}
\caption{Schematic illustration of the AquaticCLIP\textsubscript{4}. Please refer to Table 1 in the main manuscript.}
\label{fig:aquatic_clip4}
\end{figure*}

\begin{figure*}[t!]
\centering
\includegraphics[width=\linewidth]{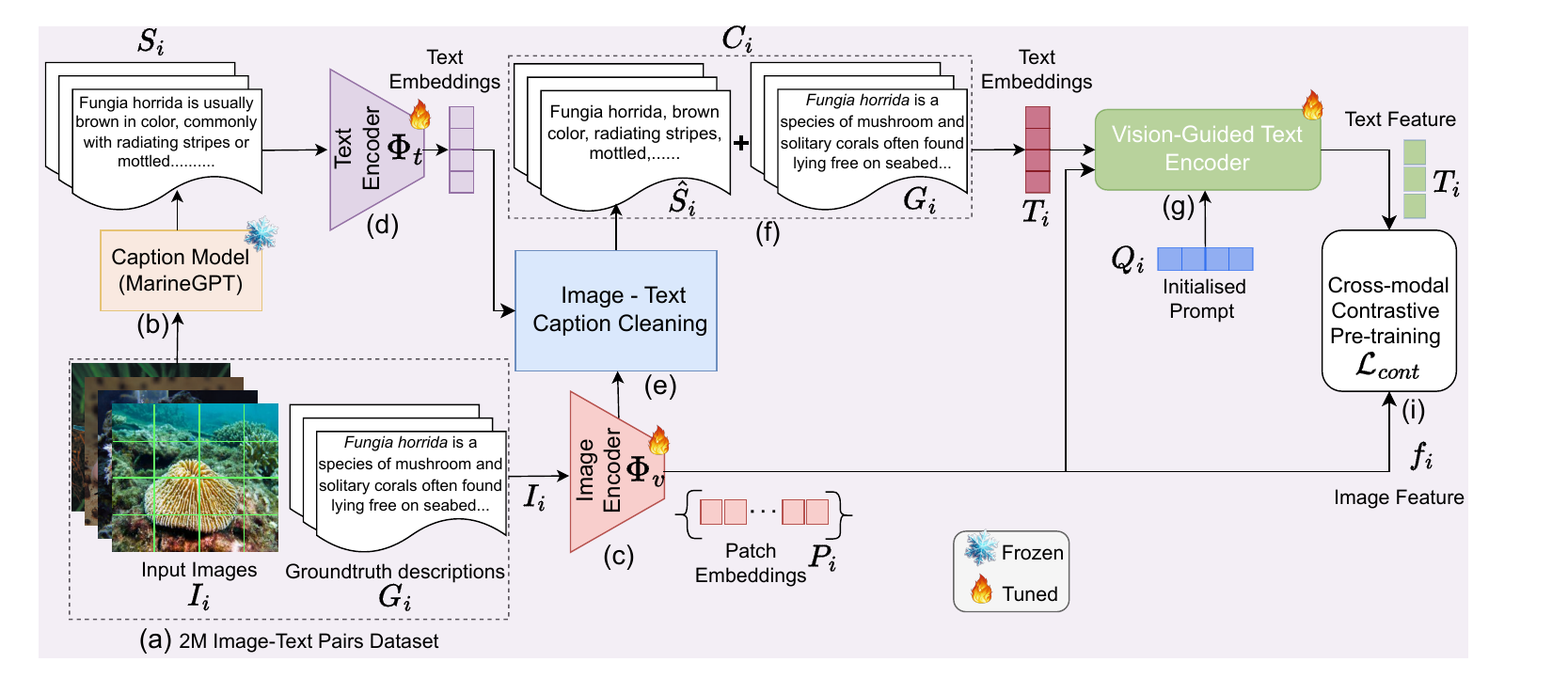}
\caption{Schematic illustration of the AquaticCLIP\textsubscript{5}. Please refer to Table 1 in the main manuscript.}
\label{fig:aquatic_clip5}
\end{figure*}

Figures \ref{fig:result_sem}-\ref{fig:result_count} display sample results from various downstream analysis tasks.

\section{Why AquaticCLIP Performance is Better?}
\label{discussion}
While MarineInst \cite{ziqiang2024marineinst} effectively performs instance segmentation and captioning in marine environments, AquaticCLIP addresses some of its limitations, primarily focusing on enhancing accuracy and addressing semantic understanding in a zero-shot setting. \\

\textbf{1. Addressing Over-Segmentation and Partial-Segmentation:} MarineInst acknowledges challenges with over-segmentation and partial-segmentation in complex marine images. 
AquaticCLIP, while not directly addressing MarineInst, tackles similar issues through its prompt-guided vision encoder (PGVE) and a dedicated textual description cleaning module (TDCM).

\begin{itemize}
\item The PGVE uses learned visual prompts to aggregate patch features, prioritizing semantically similar regions and creating a more meaningful image-level representation. 
This focus on relevant patches helps reduce the likelihood of segmenting extraneous regions, thus addressing over-segmentation.
\item  The TDCM refines the automatically generated textual descriptions by identifying and retaining the most semantically relevant keywords. This cleaning process ensures higher-quality textual descriptions, which in turn, contributes to more accurate instance identification and segmentation, potentially reducing instances of partial segmentation.
\end{itemize}
    
\noindent \textbf{2. Enhancing Semantic Understanding in a Zero-Shot Setting:} AquaticCLIP excels in zero-shot classification tasks, achieving high accuracy in recognizing marine species and coral categories not encountered during training. 
This is a significant advantage over MarineInst, which relies heavily on its pre-defined training categories for semantic understanding.

\begin{itemize}
    \item AquaticCLIP leverages the knowledge from pre-trained language models like MarineGPT, GPT4V, and BLIP2 to generate textual descriptions for the images. 
    This allows for a broader and more nuanced understanding of the visual content, even for novel classes.
     \item The use of a vision-guided text encoder (VGTE) further enhances semantic understanding by incorporating visual context from the image into the text features. 
     This alignment between visual and textual modalities results in richer semantic representations, crucial for accurate zero-shot classification.
\end{itemize}

In essence, while both models contribute significantly to marine image analysis, AquaticCLIP's focus on precise feature aggregation through PGVE, refined textual descriptions with TDCM, and enhanced semantic understanding with VGTE in a zero-shot setting addresses some limitations inherent in MarineInst's approach.

\noindent \textbf{3.AquaticCLIP's Efficient Training and Comparison to MarineInst:}
%Comparison of the dataset sizes used to train AquaticCLIP and MarineInst.
\begin{itemize}
    \item AquaticCLIP is trained on a dataset of 2 million image-text pairs. 
    This dataset focuses specifically on aquatic environments and is curated from diverse sources like YouTube, marine biology texts, online repositories, and social media etc. \
    The key is that while AquaticCLIP uses real aquatic images, the textual descriptions are pseudo-generated, meaning they are created automatically rather than manually annotated. 
    This process involves using pre-trained language models like MarineGPT and applying a cleaning module to refine the generated text.
\item MarineInst, on the other hand, is trained on a dataset called MarineInst20M, containing 2.4 million marine images with 20M object instances. 
This dataset includes images from public underwater datasets, manually collected images, and public internet images. Unlike AquaticCLIP, MarineInst20M utilizes a mixture of human-annotated and automatically generated instance masks.
\end{itemize}

Although trained on a 2M image-text paired dataset with both groundtruth and additional textual descriptions, AquaticCLIP demonstrates comparable and often superior performance to MarineInst across various marine vision tasks. 
This efficiency suggests that AquaticCLIP's architecture, particularly its use of prompt-guided vision encoding and vision-guided text encoding, effectively leverages the available data for robust and accurate analysis.

\ifCLASSOPTIONcaptionsoff
  \newpage
\fi

\bibliographystyle{IEEEtranS}
\bibliography{main}

\ifCLASSOPTIONcaptionsoff
  \newpage
\fi

\end{document}